\documentclass[twocolumn]{article}
\usepackage{graphicx}
\usepackage{titlesec}
\usepackage{amsmath,amssymb,bm,subfigure,color,url,booktabs,mathrsfs}
\usepackage[colorlinks=true,citecolor=amazonite,linkcolor=amazonite,urlcolor=amazonite]{hyperref}

\usepackage{cite}

\newcommand{\argmax}{\mathop{\rm argmax}\limits}
\definecolor{amazonite}{RGB}{0,115,150}
\definecolor{myred}{RGB}{255,56,0}
\definecolor{mygreen}{RGB}{30,150,30}
\definecolor{mybrown}{RGB}{150,30,30}
\definecolor{darkblue}{RGB}{30,50,100}
\definecolor{darkred}{RGB}{80,30,30}

\titleformat*{\section}{\color{darkred}\centering\bfseries}
\titleformat*{\subsection}{\color{darkred}\bfseries}
\titleformat*{\subsubsection}{\color{darkred}\it}
\titleformat*{\paragraph}{\color{darkred}\it}

\AtBeginDocument{
\textwidth=18.51cm
\textheight=23.5cm
\oddsidemargin=-1cm
\evensidemargin=-1cm
\headheight=0cm
\headsep=0cm
\topmargin=-0.7cm
\footskip=1cm
\columnsep=0.7cm

\heavyrulewidth=.08em
\lightrulewidth=.05em
\cmidrulewidth=.03em
\belowrulesep=.65ex
\belowbottomsep=0pt
\aboverulesep=.4ex
\abovetopsep=0pt
\cmidrulesep=\doublerulesep
\cmidrulekern=.5em
\defaultaddspace=.5em
}

\begin{document}

\twocolumn[{
\centering
{\color{darkred}\large\bf Generative Statistical Models with Self-Emergent Grammar of Chord Sequences}\\

\vskip 0.4cm
Hiroaki Tsushima$^1$, Eita Nakamura$^1$, Katsutoshi Itoyama$^1$, and Kazuyoshi Yoshii$^{1,2}$
\vskip 0.2cm

{\it\small
$^1$Graduate School of Informatics, Kyoto University, Kyoto 606-8501, Japan\\
$^2$RIKEN AIP, Tokyo 103-0027, Japan\\
}
\vskip 0.7cm
}]

\begin{abstract}
Generative statistical models of chord sequences play crucial roles in music processing. To capture syntactic similarities among certain chords (e.g.\ in C major key, between {\sf G} and {\sf G{\footnotesize 7}} and between {\sf F} and {\sf Dm}), we study hidden Markov models and probabilistic context-free grammar models with latent variables describing syntactic categories of chord symbols and their unsupervised learning techniques for inducing the latent grammar from data. Surprisingly, we find that these models often outperform conventional Markov models in predictive power, and the self-emergent categories often correspond to traditional harmonic functions. This implies the need for chord categories in harmony models from the informatics perspective.
\end{abstract}

\vspace{-2mm}
\vspace{-2mm}

\section{Introduction}
\label{sec:Intro}

It is widely accepted that harmony, or the progression of chords, plays an influential role in tonal music \cite{Rameau,Riemann,Schoenberg,Maler,TonalHarmony,BerkleeMethod}.
In the field of music informatics, computational models of chord sequences are regarded as a necessary ingredient for applications including chord recognition \cite{Papadopoulos2007,Mauch2008,Scholz2009,Lewandowski2013,McVicar2014}, harmonisation \cite{Paiement2006,Morris2008,Raczynski2013,Groves2013,Harmonisation,Whorley2016}, music analysis \cite{Winograd1968,Steedman1984,Raphael2004,Temperley2009,Kaneko2010,Rohrmeier2011,Granroth2012}, and music arrangement \cite{Fukayama2013,Quick2015,Quick2016}.
In particular, generative statistical models of chord sequences with higher predictive power are not only useful for chord sequence generation \cite{Conklin2016} and harmony analysis \cite{Raphael2004,Granroth2012} but also important for improving the accuracies of chord recognition \cite{Mauch2008,Scholz2009} and melody harmonisation \cite{Paiement2006,Raczynski2013}.
This situation parallels the case for language processing \cite{Manning,Rabiner1989,Chen1999}.

An advantage of using statistical models is that the model parameters can be learned from data.
Note that models with more precise structure do not always perform better in practice due to the overfitting problem \cite{Chen1999,Bishop}.
Thus, the performance of statistical models like predictive power should be discussed with the generalisation ability and dependence on the size of training data.
This is especially important for music modelling, for which large-size data are often not available.
For example, the Beatles dataset \cite{BeatlesData}, which is currently one of the largest datasets used for chord recognition, contains about 200 pieces.
For attempts to improve Markov models by applying smoothing techniques and/or incorporating musical knowledge, see Refs.~\cite{Papadopoulos2007,Mauch2008,Scholz2009,Yoshii2011,Raczynski2013}.

A weakness of commonly used Markov models is their limitation in learning syntactic similarities among the output symbols.
For example, in the first-order model, there is no easy way to learn tied transition probabilities $P(\,\cdot\,|x)=P(\,\cdot\,|x')$ for a pair (and in general a group) of symbols $x$ and $x'$ without knowing the pair/group a priori\footnote{The number of possible ways to tie transition probabilities is given by the partition number. For example, with 10 symbols, it is 42 for a first-order model and about $1.9\times10^8$ for a second-order model (with bigram contexts).}.
Empirically, on the other hand, we know that such similarities exist for chord symbols: e.g.\ composers often substitute, in C major key, {\sf G} with {\sf G{\footnotesize 7}} or {\sf F} with {\sf Dm} and vice versa.
This is one motivation for introducing categories of chords called harmonic functions (e.g.\ tonic, dominant, and subdominant) in harmony theories \cite{Riemann,Schoenberg,Maler,TonalHarmony,BerkleeMethod}.
If we can appropriately capture those similarities and effectively reduce model parameters, the overfitting problem would be alleviated and models with better performances could be obtained.

In generative models for natural language processing (NLP), syntactic categories like parts of speech (POS) are often represented as latent variables, e.g.\ latent states of hidden Markov models (HMMs) \cite{Johnson2007,Goldwater2007} and nonterminals of probabilistic context-free grammar (PCFG) models \cite{Johnson1998,Matsuzaki2005,Johnson}.
As the grammatical structure is described in terms of the latent variables, with a linear-chain (Markov) process in HMMs and a tree generation process in PCFG models, the syntactic similarity among words in each category is naturally captured.
Furthermore, unsupervised learning techniques can be applied to refining categories according to data \cite{Johnson1998,Matsuzaki2005} or totally inducing the latent grammar with unknown categories directly from data \cite{Johnson2007,Goldwater2007,Johnson}.
We use the terms {\it self-emergent HMMs} and {\em self-emergent PCFG models} to indicate those models whose latent variables and grammar are completely inferred through unsupervised learning using data without annotations (e.g.\ POS tags) as in the last case.
The use of these models opens the possibility of constructing generative chord sequence models incorporating syntactic categories of chords that can be adapted to the size and the style of given data.

Analogies between grammatical structures of music and language have been repeatedly discussed in the literature \cite{Winograd1968,GTTM}.
Especially, the phrase structure grammar \cite{Chomsky} has been adapted for chord sequences and harmonic functions have  been represented as nonterminals in context-free grammar (CFG) or similar tree-structured models \cite{Winograd1968,Steedman1984,Rohrmeier2011}, and more recently, in their probabilistic extensions \cite{Granroth2012,Quick2016}.
In these models, the nonterminals and production rules are manually determined according to music theories or intuitions.
Recent studies have shown that chord categories similar to traditional harmonic functions can be obtained from data by unsupervised learning \cite{Rohrmeier2008,Jacoby2015}, which implies that the generative grammar could also be inferred from data.

In this paper, we formulate self-emergent HMMs and PCFG models of chord sequences whose latent variables correspond to unknown chord categories and describe linear-chain grammar in HMMs and tree-structured grammar in PCFG models, and study unsupervised learning techniques for them.
To evaluate the predictive power and generalisation ability, we compare these models and conventionally used Markov models in terms of test-data perplexity and symbol-wise predictive accuracy for different sizes of training data.
We find that these models outperform Markov models in certain cases, typically with small training-data sizes and small numbers of possible chord symbols.
Dependence on initialisation and learning scheme is also investigated, which is known to be significant for models' performance in NLP \cite{Johnson2007,Goldwater2007,Johnson}.
We also carry out information-theoretical analyses on the structure of obtained HMMs and confirm that they indeed capture syntactic similarities among chord symbols.
Interestingly, the learned categories often correspond to traditional harmonic functions when the model size is small, which confirms the results of unsupervised chord categorisation \cite{Rohrmeier2008,Jacoby2015}.

Our main contribution is to quantitatively show the potential of self-emergent HMMs and PCFG models as generative statistical models of chord sequences, which can be applied for multiple problems in music processing.
As explained in Sec.~\ref{sec:RelatedWork}, these models have been explored in limited ways for music, and our study covers their Bayesian extensions and comparison between the expectation-maximisation (EM) algorithm and Gibbs sampling (GS) algorithm for learning.
Systematic investigations of the influence of initialisation and the relation between HMMs and PCFG models, which have practical importance, have not been discussed, to our knowledge, in the context of music modelling.
The theoretical part is intended to include a concise review of commonly used Markov models, HMMs, and PCFG models and their relations, which could be useful generally for music and language processing communities.

Our approach can be contrasted with music modelling based on expert knowledge, which has been the mainstream of music analysis.
The use of unsupervised learning removes the reliance on knowledge about functional harmony theory etc.\ and the models can be automatically adapted to the size and style of given training data.
Whereas we use chord-sequence data in the popular/jazz notation, which is of interest for applications in music processing, the method is independent of particular music styles or systems of chord notation and is applicable to other chord notations such as the Roman numeral notation and the figured bass notation.
Although we confine ourselves to information-scientific analyses in this study, we hope that the approach will be expanded to the area of computational musicology as discussed in the Conclusion.

\section{Related Work}
\label{sec:RelatedWork}

In this section, we review related studies on unsupervised learning of models for music and natural language.

\subsection{Unsupervised Learning of Music Models}
\label{sec:RelatedWorkMusic}

Unsupervised learning of chord categories has been studied in the context of music analysis.
In one study \cite{Rohrmeier2008}, chords (pitch class sets) taken from a corpus of Bach's chorales were clustered based on the hierarchical cluster analysis according to the similarity in antecedent and consequent transition probabilities of chords.
In another study \cite{Jacoby2015}, probabilistic categorisation of chord symbols was formulated based on combined optimisation of the perplexity of categories and the predictive accuracy of chords.
In both studies, chord categories similar to traditional harmonic functions were automatically obtained without referring to the pitch contents, which indicates that harmonic functions are closely related to chords' similarities in the sequential context.
In the latter study, a supplementary analysis using a self-emergent HMM with three states has been carried out and it was reported that the learned categories were not consonant with traditional harmonic functions for some data.
These studies focused on determining the optimal chord categorisation from data and did not argue whether chord categories are necessary on the ground of predictive modelling or music processing.
For example, in the formulation of the latter study \cite{Jacoby2015}, any categorisation scheme cannot exceed Markov models in the predictive power.

In Ref.~\cite{Quick2016}, unsupervised learning was applied to a variant of PCFG models for chord symbols in which nonterminals describe harmonic functions.
In this study, the nonterminals and the production rules were manually determined in accordance with existing data of harmony analysis and only the probabilities were learned unsupervisedly.
Our PCFG models can be considered as extensions of this model as we allow larger numbers of nonterminals and they have no prior labels or restricted production rules.

Unsupervised learning of self-emergent HMMs have been applied for analyses of melodies \cite{Mavromatis2009,Mavromatis2012}, which demonstrated that learned HMM states can capture the syntactic roles of musical symbols such as pitch and note value.
In the first study \cite{Mavromatis2009}, pitch sequences of Greek church chants were modelled with HMMs and it was found that the learned HMM structures can be related to melodic cadential formulas.
In the second study \cite{Mavromatis2012}, sequences of note values extracted from Palestrina's masses were analysed similarly and it was found that the learned HMM states corresponded to metric positions of note onsets.
Although the EM algorithm was used for learning, whose results generally depend on initial parameters, the influence of initialisation was not discussed.
The optimal number of latent states was estimated by the minimum description length principle, but the dependence on the training-data size was not studied and there was no comparison with Markov models or other models in terms of perplexity etc.

In the context of melody modelling, comparison of Markov models, HMMs, and PCFG models have been carried out \cite{Abdallah2014}.
Zeroth- and first-order Markov models of pitches and of pitch intervals, self-emergent HMMs with up to 12 latent states, and PCFG models with pre-assigned rules were formulated as Bayesian models and compared in terms of variational free energy.
The results showed that the relative performance of the models depended on the data: for Bach's chorales a PCFG model was the best and for a collection of folk songs the first-order Markov model of pitch intervals was the best.
Because higher-order Markov models or larger-size HMMs were not tested and models were learned and tested using the same data, the overfitting effect could not be observed.
The influence of initialisation was also not discussed.

\subsection{Unsupervised Learning of Language Models}

Self-emergent HMMs have been applied for unsupervised POS tagging \cite{Johnson2007,Goldwater2007}.
In Ref.~\cite{Goldwater2007}, Bayesian HMMs were learned using GS and they were shown to yield significantly higher tagging accuracies than HMMs trained by the EM algorithm.
In Ref.~\cite{Johnson2007}, it was shown that the performance of HMMs learned by the EM algorithm is largely influenced by the random initialisation and when the size of the state space is small (e.g.\ 25), they yielded a similar tagging accuracy as the Bayesian HMMs.
Since the tagging accuracy was the focus of these studies, the perplexities of the output symbols (words) were not measured.

Unsupervised learning of PCFG models was studied in the context of grammar induction and parsing \cite{Johnson}.
The maximum likelihood estimation using the EM algorithm and Bayesian estimation using GS were compared and the latter method was shown to give higher accuracies in some cases.
It was argued that even though for both learning schemes fully unsupervised grammar induction using simple PCFG models is difficult, Bayesian learning seems to induce more meaningful grammar owing to its ability to prefer sparse grammars.

Unsupervised learning has also been applied to adapt statistical language models for particular data.
For NLP, there exist corpora of syntactic trees that are based on widely accepted linguistic theories (e.g.\ \cite{Chomsky}), from which production probabilities can be obtained by supervised learning.
On the other hand, studies have shown that extending the standard annotation symbols (e.g.\ NP) to incorporate context dependence and subcategories of words, often called symbol refinement, improves the accuracy of parsing \cite{Johnson1998}.
Unsupervised learning of variants of PCFG models has been successfully applied to find optimal refinement of nonterminal symbols from data \cite{Matsuzaki2005}.

\section{Models}
\label{sec:Model}

Here we describe Markov models and self-emergent HMMs and PCFG models for chord symbols and their learning methods.
The problem of initialisation for unsupervised learning of self-emergent HMMs and PCFG models is discussed and two learning methods, the expectation-maximisation (EM) algorithm and Gibbs sampling (GS) method, are explained.

A sequence of chord symbols is written as
\begin{equation}
\bm x=x_{1:N}=(x_1,\ldots,x_N)~,
\label{eq:dataSeq}
\end{equation}
where each $x_n$ takes values in the set of chord symbols (output symbol space) $\Omega=\{x\}$.
The number of possible symbols (symbol-space size) is denoted by $N_\Omega=\#\Omega$.
In this study, a generative statistical model of chord sequences is defined as a model that yields the probability $P(x_{1:N})$ for any sequence $x_{1:N}$.

\subsection{Markov Model}
\label{sec:MarkovModel}

The first-order Markov model is defined with an initial probability ($\chi^{\rm ini})$ and transition probabilities ($\chi$) as
\begin{equation}
P(x_1)=\chi^{\rm ini}_{x_1},\quad P(x_n|x_{n-1})=\chi_{x_{n-1}x_n}.
\end{equation}
The probability of $x_{1:N}$ is then given as
\begin{equation}
P(x_{1:N})=\chi^{\rm ini}_{x_1}\prod_{n=2}^N\chi_{x_n x_{n-1}}.
\label{eq:MarkovModelProb}
\end{equation}
As a generalisation, a $k$\,th-order Markov model is defined with the following initial and transition probabilities:
\begin{align}
P(x_1)&=\chi^{\rm ini}_{x_1},
\\
P(x_2|x_1)&=\chi^{{\rm ini}(2)}_{x_1x_2},
\\
&~\;\vdots
\notag
\\
P(x_k|\,x_1,\ldots,x_{k-1})&=\chi^{{\rm ini}(k)}_{x_1\cdots x_{k-1}x_k},
\\
P(x_n|\,x_{n-k},\ldots,x_{n-1})&=\chi_{x_{n-k}\cdots x_{n-1}x_n}.
\end{align}
The probability $P(x_{1:N})$ is given similarly as in Eq.~(\ref{eq:MarkovModelProb}).
Higher-order Markov models are more precise than lower-order Markov models.
This is because for any $k$\,th-order Markov model we can choose parameters for a $(k{+}1)$th-order Markov model so that these models yield the same probability $P(x_{1:N})$ for any sequence $x_{1:N}$.

Since each $x_n$ can take $N_\Omega$ values, for the first-order model, the initial probability has $N_\Omega$ parameters and the transition probability has $N_\Omega^2$ parameters.
These parameters are not independent, however, since the probabilities must add up to unity.
After normalisation, we have $N_\Omega-1$ free parameters for the initial probability and $N_\Omega(N_\Omega-1)$ for the transition probability, and hence $(1+N_\Omega)(N_\Omega-1)$ free parameters in total.
Similar calculations yield the number of free parameters for a $k$\,th-order Markov model as in Table \ref{tab:Parameters}.

Given a training dataset, the parameters of Markov models can be learned based on the maximum likelihood principle.
For example, the transition probability $P(x'|x)$ for the first-order Markov model is estimated as
\begin{equation}
P(x'|x)\propto C(xx'),
\label{eq:MLTrProb}
\end{equation}
where $C(xx')$ denotes the number of times that a bigram $xx'$ appears in the training data.
Other parameters can be obtained similarly.
Especially for small training data and for high-order Markov models, however, smoothing techniques are necessary to avoid the data sparseness problem.
For the additive smoothing, we simply add a positive constant number $\epsilon$ to the right-hand side (RHS) of Eq.~(\ref{eq:MLTrProb}) and similarly for other cases.
Based on a previous study that compared various smoothing methods for language data, the most effective smoothing method was the (interpolated) Knesner-Ney (KN) smoothing and the modified KN (MKN) smoothing, which incrementally use lower-order transition probabilities to obtain higher-order transition probabilities \cite{Chen1999}.
\begin{table}[t]
\centering
{\tabcolsep = 2pt
\begin{tabular}{lr}\toprule
Model & Number of free parameters\\
\midrule
Markov & $(1+N_\Omega+\cdots+N_\Omega^k)(N_\Omega-1)$\\
HMM    & $(1+N_\Gamma)(N_\Gamma-1)+N_\Gamma(N_\Omega-1)$\\
PCFG   & $(1+N_\Delta)(N_\Delta^2-1)+N_\Delta N_\Omega$\\
\bottomrule
\end{tabular}
}
\caption{Number of model parameters after normalisation for a $k$\,th order Markov model, an HMM with $N_\Gamma$ states, and a PCFG model with $N_\Delta$ nonterminals and the constraint $\psi_{S\to x}\equiv0$. $N_\Omega$ is the symbol-space size.}
\label{tab:Parameters}
\vspace{-2mm}
\end{table}
%

\subsection{HMM}

\subsubsection{Basic Model and the EM Algorithm}
\label{sec:BasicHMM}

In HMMs, we introduce a latent variable (state), denoted by $z_n$, corresponding to each symbol $x_n$ and consider a first-order Markov model for the sequence of latent states.
Let $\Gamma=\{z\}$ denote the set of possible latent states called the state space and $N_\Gamma=\#\Gamma$ denote its size.
The initial and transition probabilities will be denoted by $\pi^{\rm ini}$ and $\pi$.
In addition, each $x_n$ is considered to be drawn from a distribution depending on $z_n$ called the output probability ($\phi$).
Thus, the model parameters are as follows:
\begin{align}
P(z_1)=\pi^{\rm ini}_{z_1},\quad P(&z_n|z_{n-1})=\pi_{z_{n-1}z_n},
\label{eq:HMMTrProb}
\\
P(x_n|z_n)&= \phi_{z_nx_n}.
\label{eq:HMMOutProb}
\end{align}
The number of parameters is as shown in Table \ref{tab:Parameters}.
We interpret the latent states as chord categories.
The transition probabilities describe the linear-chain grammar for chord categories and the output probabilities relate the categories to chord symbols.

We can also define higher-order HMMs by considering higher-order Markov models for latent states.
For self-emergent HMMs (i.e.\ if the state space is not fixed), however, first-order HMMs are general enough that higher-order HMMs can be described by using them.
To see this, consider a second-order HMM with a state space $\Gamma=\{z\}$ and transition probability $P(z_n|z_{n-1},z_{n-2})$.
If we define a state space $\hat{\Gamma}$ as the product space $\Gamma\times\Gamma$ and identify $\hat{z}_n\in\hat{\Gamma}$ as $(z_n,z_{n-1})\in\Gamma\times\Gamma$, the transition probability $P(\hat{z}_n|\hat{z}_{n-1})$ is equivalent to $P(z_n|z_{n-1},z_{n-2})$ since
\begin{equation}
P(z_n,z_{n-1}\,|\,z_{n-1},z_{n-2})=P(z_n|\,z_{n-1},z_{n-2}).
\end{equation}
A similar argument is valid for the initial and output probabilities and for higher-order HMMs.

In the unsupervised case, i.e.\ when the training data only has the information of output symbols but not corresponding latent states, the parameters of HMMs cannot be learned as simply as the case for Markov models.
Nevertheless, the EM algorithm can be applied for the maximum likelihood estimation.
The EM algorithm (also called the Baum-Welch algorithm) is based on iterative optimisation.
Given training data $\bar{X}=x_{1:\bar{N}}$ (for notational simplicity, we here consider that the training data has only one sequence; the extension for data with multiple sequences is straightforward) and a set of parameters $\Pi=(\pi^{\rm ini}_z,\pi_{zz'},\phi_{zx})$, the updated parameters $\tilde{\Pi}=(\tilde{\pi}^{\rm ini}_z,\tilde{\pi}_{zz'},\tilde{\phi}_{zx})$ are given by
\begin{align}
\tilde{\pi}^{\rm ini}_z&=P(z_1=z\,|\,\bar{X},\Pi),
\\
\tilde{\pi}_{zz'}&=\frac{\sum\limits_{n=1}^{\bar{N}-1}P(z_n=z,z_{n+1}=z'\,|\,\bar{X},\Pi)}{\sum\limits_{n=1}^{\bar{N}-1}P(z_n=z\,|\,\bar{X},\Pi)},
\\
\tilde{\phi}_{zx}&=\frac{\sum\limits_{n=1}^{\bar{N}}\delta_{x_nx}P(z_n=z\,|\,\bar{X},\Pi)}{\sum\limits_{n=1}^{\bar{N}}P(z_n=z\,|\,\bar{X},\Pi)}.
\end{align}
Here, the probabilities in the RHS are defined with the parameters before the update and can be computed by the forward-backward algorithm (see e.g.\ Ref.~\cite{Rabiner1989} for details).
These updates always increase the likelihood, $P(\bar{X}|\tilde{\Pi})\geq P(\bar{X}|\Pi)$, until it converges to a local maximum \cite{Bishop}.
According to this monotonic increase of the likelihood, the EM algorithm can also get stuck in a local maximum and miss the global maximum.

There is no guarantee that the EM algorithm finds the global maximum likelihood solution.
Thus we have the problem of initialisation for unsupervised learning of HMMs.
Possible approaches are (i) to start with several random initial values and obtain the result with the maximum likelihood and (ii) to start with a potentially `good' choice of parameters, e.g.\ parameters estimated by supervised learning with small-size data; parameters set by existing theories, intuition, etc.

\subsubsection{Bayesian Model and the GS Method}

Bayesian learning can be applied for learning HMM parameters to alleviate the potential problems of the EM algorithm \cite{Johnson2007,Goldwater2007}.
Firstly, with the introduction of the prior distribution of the parameters, one can incorporate prior knowledge or preferences into parameter learning.
Secondly and more importantly for us, there are sampling-based algorithms that can search parameters without getting stuck in local optima.

To utilise Bayesian learning techniques, we extend the model with prior distributions on the parameters.
Since the initial, transition, and output probabilities are all categorical distributions, we put the conjugate Dirichlet distributions on their parameters $\bm\pi^{\rm ini}=(\pi^{\rm ini}_z)_z$, $\bm\pi_z=(\pi^{\rm ini}_{zw})_w$, and $\bm\phi_z=(\phi_{zx})_x$ as
\begin{align}
\bm\pi^{\rm ini}\sim{\rm Dir}(\bm\lambda^{\rm ini}),\quad\bm\pi_z\sim{\rm Dir}(\bm\lambda_z),\quad\bm\phi_z\sim\;&{\rm Dir}(\bm\rho_z),
\end{align}
where $\bm\lambda^{\rm ini}$, $\bm\lambda_z$, and $\bm\rho_z$ are Dirichlet parameters with dimensions equal to the corresponding distributions.

For Bayesian inference, we estimate the distribution of the parameters $P(\Pi|\,\bar{X},\Lambda)$ given the training data $\bar{X}$ and hyperparameters $\Lambda=(\bm\lambda^{\rm ini},\bm\lambda_z,\bm\rho_z)$.
The Gibbs sampling (GS) is a method to sample $Z=z_{1:\bar{N}}$ and $\Pi$ from the joint distribution $P(Z,\Pi\,|\,\bar{X},\Lambda)$.
The variables are alternately sampled from the distributions $P(Z|\,\Pi,\bar{X},\Lambda)$ and $P(\Pi|\,Z,\bar{X},\Lambda)$ \cite{Goldwater2007}.
In theory, the Gibbs sampler yields samples from the exact distribution $P(\Pi|\,\bar{X},\Lambda)$.
This means that the sampled parameters explore globally in the parameter space with more weight on the regions with larger likelihood.
In practice, we select the maximum likelihood parameters out of samples from the Gibbs sampler.
Because the sampled parameters are not always locally optimal, we apply the EM algorithm using the selected sample as initial values, to finally obtain locally optimal parameters.

\subsection{PCFG Model}

\subsubsection{Basic Model and the EM Algorithm}

PCFG models are defined with a set of output symbols (also called terminals) $\Omega=\{x\}$, a set of latent variables called nonterminals $\Delta=\{z\}$ (its size is denoted by $N_\Delta$), a start symbol $S$, and the following set of production probabilities and output probabilities\footnote{To make the relation between PCFG models and HMMs explicit, we here do not include the start symbol in $\Delta$.}:
\begin{align}
P(S\to z_Lz_R)&=\theta_{S\to z_Lz_R},\quad
P(S\to x)=\psi_{S\to x},
\\
P(z\to z_Lz_R)&=\theta_{z\to z_Lz_R},\quad
P(z\to x)=\psi_{z\to x}.
\end{align}
As the case for HMMs, we interpret the nonterminals as chord categories.
The production probabilities ($\theta$) describe the tree-structured grammar for chord categories and the output probabilities ($\psi$) relate the categories to chord symbols.
The probabilities are normalised as
\begin{align}
1&=\sum_{z_L,z_R\in\Delta}\theta_{S\to z_Lz_R}+\sum_{x\in\Omega}\psi_{S\to x},
\\
1&=\sum_{z_L,z_R\in\Delta}\theta_{z\to z_Lz_R}+\sum_{x\in\Omega}\psi_{z\to x}.
\end{align}
Notice that the rule $S\to x$ is only used for sequences of length unity.
Whereas we retain this probability in this section, which will be used for theoretical discussions, we set $\psi_{S\to x}=0$ practically for numerical calculations because there were no chord sequences with length unity in our datasets.
With this constraint, the number of parameters is as shown in Table \ref{tab:Parameters}.

We can derive the EM algorithm for PCFG models similarly as the case for HMMs.
Given training data $\bar{X}$ and a set of parameters $\Theta=(\theta_{S\to z_Lz_R},\theta_{z\to z_Lz_R},\psi_{S\to x},\psi_{z\to x})$, the updated parameters {\color{white}\Big|\!}$\tilde{\Theta}=\big(\tilde{\theta}_{S\to z_Lz_R},\tilde{\theta}_{z\to z_Lz_R},\tilde{\psi}_{S\to x},\tilde{\psi}_{z\to x}\big)$ are given as
\begin{equation}
\tilde{\theta}_{z\to z_Lz_R}\propto\sum_{T\in\mathbb{T}(\bar{X})}c(z\to z_Lz_R;T)P(T|\,\bar{X},\Theta)
\label{eq:PCFGEMUpdate}
\end{equation}
and similarly for $\tilde{\theta}_{S\to z_Lz_R}$, $\tilde{\psi}_{S\to x}$, and $\tilde{\psi}_{z\to x}$.
Here, $T$ denotes a derivation tree, $\mathbb{T}(\bar{X})$ is the set of possible derivation trees that yield $\bar{X}$, and $c(R;T)$ is the number of times that a production rule $R$ is used in $T$.
The probability $P(T|\,\bar{X},\Theta)$ can be expanded as
\begin{align}
P(T|\,\bar{X},\Theta)=
&\prod_{z_L,z_R}\theta_{S\to z_Lz_R}^{c(S\to z_Lz_R;T)}\prod_{x}\psi_{S\to x}^{c(S\to x;T)}
\notag
\\
&\cdot\prod_{z,z_L,z_R}\theta_{z\to z_Lz_R}^{c(z\to z_Lz_R;T)}\prod_{z,x}\psi_{z\to x}^{c(z\to x;T)}
\notag
\end{align}
and the RHS of Eq.~(\ref{eq:PCFGEMUpdate}) can be computed by the inside-outside algorithm \cite{Manning}.
The discussion about initialisation for HMMs is also valid for PCFG models.

\subsubsection{Bayesian Model and the GS Method}

The Bayesian extension for PCFG models is also similar as the case for HMMs.
We put conjugate priors for the model parameters as
\begin{align}
\bm\theta_S\sim{\rm Dir}(\bm\xi_S),\quad\bm\psi_S\sim{\rm Dir}(\bm\zeta_S),
\\
\bm\theta_z\sim{\rm Dir}(\bm\xi_z),\quad\bm\psi_z\sim{\rm Dir}(\bm\zeta_z),
\end{align}
where $\bm\theta_S=(\theta_{S\to z_Lz_R})_{z_Lz_R}$, $\bm\theta_z=(\theta_{z\to z_Lz_R})_{z_L,z_R}$, $\bm\psi_S=(\psi_{S\to x})_x$, and $\bm\psi_z=(\psi_{z\to x})_x$ are the vector representation of the probabilities and $\bm\xi_S$, $\bm\xi_z$, $\bm\zeta_S$, and $\bm\zeta_z$ are Dirichlet parameters, with dimension $N_\Delta^2$, $N_\Delta^2$, $N_\Omega$, and $N_\Omega$.

In the GS for PCFG models, we alternately sample derivation trees and model parameters.
Let us denote the hyperparameters as $\Xi=(\bm\xi_S,\bm\xi_z,\bm\zeta_S,\bm\zeta_z)$.
The variables are sampled from $P(T|\,\Theta,\bar{X},\Xi)$ and $P(\Theta|\,T,\bar{X},\Xi)$, respectively.
The derivation trees can be sampled by the inside-filtering outside-sampling method and the model parameters can be sampled from the posterior Dirichlet distributions in a standard way \cite{Johnson}.

\section{Model Analysis Techniques}
\label{sec:TheorAnalyses}

Methods for evaluating and analysing the learned models are developed here.
In Sec.~\ref{sec:ModelRelations}, we discuss theoretical relations among the three types of models (Markov models, HMMs, and PCFG models) and derive an initialisation method for PCFG models using parameters of corresponding HMMs.
In Sec.~\ref{sec:EvaluationMeasure}, we formulate several evaluation metrics to measure the predictive power of the generative models.
In Sec.~\ref{sec:HMMAnalysis}, we develop techniques for examining the structure of self-emergent HMMs, which will be used in Sec.~\ref{sec:InfTheorAnalysis} to confirm that those models can indeed capture syntactic similarities among chord symbols.

\subsection{Relations among the Models}
\label{sec:ModelRelations}

We here explain theoretical relations among Markov models, HMMs, and PCFG models, which provide insights into the model analysis.
We use the expression that a model (say model {\sf A}) can be {\it mimicked} by another model (say model {\sf B}) if for all possible parameter configurations for model {\sf A} there is always a parameter configuration for model {\sf B} that provides the same probability $P(X;{\sf A})=P(X;{\sf B})$ for all data $X$.
We also say that a set of models {\it includes} another set of models if any model in the latter set can be mimicked by a model in the former set.
In summary, we show that smaller HMMs/PCFG models with less number of latent states or nonterminals can be mimicked by larger HMMs/PCFG models.
In addition, HMMs include Markov models and PCFG models include HMMs.

The first assertion is easy to see.
The parameters for an HMM with $N_\Gamma$ states can be extended to those for a new HMM with $N_\Gamma{+}1$ states by assigning $\pi^{\rm ini}_{\hat{z}}=\pi_{z\hat{z}}=0$ for the newly added state $\hat{z}$.
In order to complete the new HMM, $\pi_{\hat{z}z}$ and $\phi_{\hat{z}x}$ must be specified, but these parameters do not affect the probability $P(\bm x|\Pi)$ for any sequence $\bm x$ since $\hat{z}$ will be generated with zero probability.
The same argument is valid for PCFG models.

\begin{figure}[t]
\centering
\includegraphics[clip,width=0.6\columnwidth]{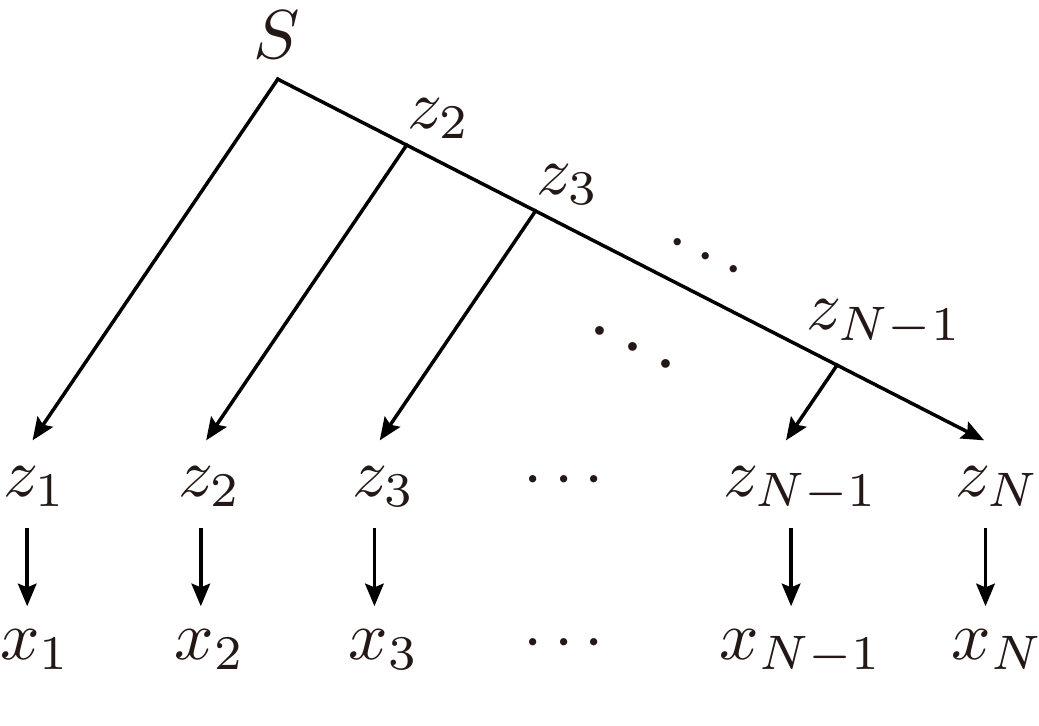}
\vspace{-5mm}
\caption{Realisation of linear-chain grammar in the tree-structured grammar.}
\label{fig:HMMEmbedding}
\vspace{-3mm}
\end{figure}
Next, to see that HMMs include Markov models, we can construct an HMM that mimics any given Markov model.
For any first-order Markov model, we can indeed construct such an HMM by setting $\Gamma=\Omega$, $\pi^{\rm ini}=\chi^{\rm ini}$, $\pi=\chi$, and $\phi_{zx}=\delta_{zx}$.
Since any higher-order Markov model can be represented as a first-order Markov model with an extended set of output symbols, as we saw in the discussion for higher-order HMMs, this shows that the set of HMMs includes Markov models of any order.

The relation between HMMs and PCFG models derives from a fact that a linear-chain grammar can be realised as a special case of tree-structured grammar as illustrated in Fig.~\ref{fig:HMMEmbedding}.
Roughly speaking, for a given HMM with $N_\Gamma$ states, we can define a PCFG model with $N_\Delta=N_\Gamma$ nonterminals with the following probabilities
\begin{align}
\theta_{S\to z_Lz_R}&=\pi^{\rm ini}_{z_L}\pi_{z_Lz_R}(1-\kappa'),
\label{eq:PCFGMimickingHMM1}
\\
\psi_{S\to x}&=\sum_z\pi^{\rm ini}_{z}\phi_{zx}\kappa',
\label{eq:PCFGMimickingHMM2}
\\
\theta_{z\to z_Lz_R}&=\delta_{zz_L}\pi_{zz_R}(1-\kappa),
\label{eq:PCFGMimickingHMM3}
\\
\psi_{z\to x}&=\phi_{zx}\kappa
\label{eq:PCFGMimickingHMM4}
\end{align}
with some constants $0\leq\kappa,\kappa'\leq1$, so that the derivation tree in Fig.~\ref{fig:HMMEmbedding} yields the probability (for $N>1$)
\begin{align}
P(x_{1:N},z_{1:N}|\Theta)= {}&\kappa^N(1-\kappa)^{N-2}(1-\kappa')
\notag
\\
&\cdot\pi^{\rm ini}_{z_1}\phi_{z_1x_1}\prod_{n=2}^N(\pi_{z_{n-1}z_n}\phi_{z_nx_n}),
\end{align}
which reproduces the complete-data probability of the HMM up to a constant factor involving $\kappa$ and $\kappa'$.
To show a strict inclusion, however, we need more careful derivation because of the different normalisation conditions for HMMs and PCFG models.
In fact, the inclusion relation is strictly true only when HMMs and PCFG models are appropriately reformulated as we show in Appendix \ref{app:HMMEmbedding}.
For here, we remark that the inclusion is not strictly true with our definition of HMMs, since the probabilities in Eqs.~(\ref{eq:PCFGMimickingHMM1})--(\ref{eq:PCFGMimickingHMM4}) can generate derivation trees with different structures.
It is also remarked that $\kappa'$ controls the probability of generating sequences of length unity.
Since all chord sequences used in our study have lengths larger than unity, we exclude this case and consider $\kappa'=0$.

The inclusion relation between HMMs and PCFG models can be applied to finding initial values for parameter learning of PCFG models \cite{Lari1990}.
It has been reported that the initialisation problem is more severe for PCFG models than HMMs because of the complex model structure and the existence of many local optima \cite{Johnson}.
Thus, if we were given parameters of HMMs, the initialisation problem could be alleviated if we use them to set initial values for PCFG models as in Eqs.~(\ref{eq:PCFGMimickingHMM1})--(\ref{eq:PCFGMimickingHMM4}).
Several remarks should be made here.
First, we need to specify the constant $\kappa$.
As we have $\sum_x\psi_{z\to x}=\kappa$ from Eq.~(\ref{eq:PCFGMimickingHMM4}), this parameter describes the probability that a nonterminal is converted to an output symbol (rather than being split into two nonterminals), which affects the expected length $\langle N\rangle$ of generated sequences.
Indeed, we can derive the following equation (valid for $\kappa>1/2$ and $\kappa'=0$):
\begin{equation}
\langle N\rangle=\frac{2\kappa}{2\kappa-1},
\label{eq:ExpectedLength}
\end{equation}
which can be used to choose $\kappa$.
Second, since the constraint $z=z_L$ in Eq.~(\ref{eq:PCFGMimickingHMM3}) cannot be relaxed during the learning process (for both EM and GS algorithms) once imposed, we should modify the equation to increase the flexibility of learning.
For this, we can add a small positive constant $\eta$ to the RHS of Eq.~(\ref{eq:PCFGMimickingHMM3}) and renormalise the probability.

\subsection{Evaluation Measures}
\label{sec:EvaluationMeasure}

As a measure of evaluating statistical models, the test-data perplexity (equivalent to the cross entropy) is commonly used.
Denoting the test data by $X$ and the learned parameters by $\bar{\Theta}$, the {\it test-data perplexity} ${\cal P}$ is given as
\begin{equation}
{\cal P}={\rm exp}\Big(-\frac{1}{|X|}{\rm ln}\,P(X|\bar{\Theta})\Big),
\end{equation}
where $|X|$ denotes the number of samples in $X$ and $P(X|\bar{\Theta})$ is called the {\it evidence probability}.
As noted above, we need to use the equivalent normalisation condition to compare PCFG models with other models appropriately.
Calculation of the normalised evidence probability for PCFG models is explained in Appendix \ref{app:NormalisedEvidenceProb}.

The test-data perplexity is (the exponential of) the entropy rate of the test data measured with the learned model parameters and defined with the evidence probability for the whole symbol sequences.
While it is a measure of model's predictive power in a general sense, this measure is defined without making explicit predictions.
We can also consider quantities that measure how the learned model can predict each symbol from the rest of the symbols in a sequence.
To define these, we first calculate the probability
\begin{equation}
P(x_n=x\,|\bm x_{\neg n})=P(x_n=x\,|\,x_{1:n-1},x_{n+1:N})
\label{eq:SymbolPredictionProb}
\end{equation}
for each symbol $x_n$ in each sequence $\bm x=x_{1:N}$ in $X$.
Here, $\bm x_{\neg n}$ denotes the sequence obtained from $\bm x$ by removing $x_n$ and the dependence on $\bar{\Theta}$ is made implicit for simplicity.
We can choose the symbol that maximises this probability as a prediction for $x_n$ and the {\it symbol-wise predictive error rate} (or simply, {\it error rate}) ${\cal E}$ can be defined as
\begin{equation}
{\cal E}=\frac{1}{|X|}\sum_{\bm x\in X}\sum_{x_n\in\bm x}\mathbb{I}(x_n\neq\argmax_xP(x|\bm x_{\neg n})),
\end{equation}
where $\mathbb{I}({\cal C})$ is $1$ if condition ${\cal C}$ is true and 0 otherwise.
A potential problem with the error rate is that it cannot take into account how the prediction is close to the ground truth if it is incorrect.
An alternative measure can be defined by ranking the multiple predictions and `averaging' the ranks of the ground-truth symbol.
According to the probability in Eq.~(\ref{eq:SymbolPredictionProb}), we can sort the candidate predictions and calculate the rank $r(x_n)$ of the ground-truth symbol.
We define the {\it reciprocal of the mean reciprocal rank} (RMRR; the harmonic mean of the ranks) as
\begin{equation}
{\cal R}=\bigg[\frac{1}{|X|}\sum_{\bm x\in X}\sum_{x_n\in\bm x}\frac{1}{r(x_n)}\bigg]^{-1},
\label{eq:RMRR}
\end{equation}
which is commonly used for evaluating classification methods.
Formulas for calculating Eq.~(\ref{eq:SymbolPredictionProb}) for the three types of models we consider are derived in Appendix \ref{app:SymbolPrediction}.

\subsection{Information-Theoretical Analysis of HMMs}
\label{sec:HMMAnalysis}

As we will see in Sec.~\ref{sec:Experiment}, HMMs often perform better than Markov models in terms of predictive power.
Here we develop analysis techniques to examine how the structure of self-emergent HMMs learned from data evolves with increasing sizes of the state space.

In principle, when the model size is increased, HMMs can evolve in two ways to increase their predictive power (or decrease the perplexity).
One way is to give sparser output probabilities for each state.
Since the output probabilities determine the resolution of assigning symbols to each state, their sparseness contributes to accurate prediction of symbols.
The extreme case is achieved when output probabilities have all zero probabilities except for one symbol as in Markov models.
The other way is to associate multiple states to one or a group of symbol(s) and describe finer sequential dependence with distinct transition probabilities.
In this way transition probabilities can incorporate dependence on more precise contexts including longer-range contexts, similarly as a higher-order HMM is represented as a first-order HMM.
Since these two directions are in a trade-off relation when the state space is finite, how learned HMMs actually behave depends on (or characterises) the nature of used data.
Thus, although the following methods are model analysis techniques, they can also reveal the nature of chord sequence data from the information-theoretical perspective.

Let us now derive quantities that characterise the structure of HMMs.
We first introduce the stationary distribution of latent states $\hat{P}(z)$, which is defined by the equilibrium equation as
\begin{equation}
\hat{P}(z)=\sum_{w\in\Gamma}\hat{P}(w)\pi_{wz}.
\label{eq:StationaryProb}
\end{equation}
(Recall the notation for HMMs in Sec.~\ref{sec:BasicHMM}: $z$ and $w$ denote latent states, $\Gamma$ the latent state space, and $\pi$ the transition probability.)
Because of the ergodic property of HMMs, the stationary distribution can be seen as the unigram probability of latent states generated from the HMM, and we use the notation $\hat{P}$ to indicate that the probability is considered `locally' by ignoring the sequential dependencies.
The first quantity we define is the {\it perplexity of the stationary distribution} ${\cal P}_\Gamma$ given as
\begin{equation}
{\cal P}_\Gamma={\rm exp}\bigg[-\sum_{z\in\Gamma}\hat{P}(z)\,{\rm ln}\,\hat{P}(z)\bigg].
\label{eq:StationaryPerp}
\end{equation}
This quantity is interpreted as the effective size of the state space that is used to describe data.

The sparseness of the output probabilities can be measured by means of their entropy.
We define the {\it average perplexity of output probabilities} ${\cal P}_\phi$ as
\begin{equation}
{\cal P}_\phi={\rm exp}\bigg[-\sum_{z\in\Gamma}\hat{P}(z)\sum_{x\in\Omega}\phi_{zx}\,{\rm ln}\,\phi_{zx}\bigg].
\end{equation}
(Recall that $x$ denotes a chord symbol and $\Omega$ the set of possible chord symbols.)
Intuitively, this quantity tells how many symbols are assigned to each state on average.
Another quantity can be obtained by considering the variety of association between symbols and states in the opposite way.
The unigram probability of symbols generated by the HMM $\hat{P}(x)$ and the probability of states given a symbol $\hat{P}(z|x)$ can be expressed as follows:
\begin{align}
\hat{P}(x)&=\sum_{z\in\Gamma} \hat{P}(z)\phi_{zx},
\\
\hat{P}(z|x)&=\frac{\hat{P}(z,x)}{\hat{P}(x)}=\frac{\hat{P}(z)\phi_{zx}}{\hat{P}(x)}.
\end{align}
The {\it average state association variety} ${\cal V}$ is defined by first calculating the entropy of $\hat{P}(z|x)$ with respect to $z$ for each $x$ and then averaging them with $\hat{P}(x)$ (and finally exponentiating):
\begin{equation}
{\cal V}={\rm exp}\bigg[-\sum_{x\in\Omega} \hat{P}(x)\sum_{z\in\Gamma} \hat{P}(z|x)\,{\rm ln}\,\hat{P}(z|x)\bigg].
\end{equation}
This quantity describes how many states are associated to each symbol effectively.

The last quantity we define is the {\it average perplexity of transition probabilities} ${\cal P}_{\pi}$ given as
\begin{equation}
{\cal P}_{\pi}={\rm exp}\bigg[-\sum_{z\in\Gamma}\hat{P}(z)\sum_{w\in\Gamma}\pi_{zw}\,{\rm ln}\,\pi_{zw}\bigg].
\end{equation}
This quantity measures the average sparseness of the transition probabilities, or the effective number of possible states that can be reached from one state.
Low perplexities of transition probabilities are necessary for accurate prediction of latent states and thence that of symbols.
This quantity should be considered in relation with the stationary perplexity ${\cal P}_\Gamma$ in Eq.~(\ref{eq:StationaryPerp}), which ignores the sequential dependence of state transitions.

\section{Numerical Experiments}
\label{sec:Experiment}

Here we present numerical results.
After describing the data and the setup in Sec.~\ref{sec:DataAndSetup}, the results of model comparisons are presented in Secs.~\ref{sec:MarkovVSHMM} and \ref{sec:PCFGVSHMM}, where the influence of initialisation for unsupervised learning is also discussed.
In Sec.~\ref{sec:StructureOfSelfEmergentGrammar}, we examine the structure of self-emergent grammar of learned models.
Implications of the results are discussed in Sec.~\ref{sec:Discussion}.

\subsection{Datasets and Setup}
\label{sec:DataAndSetup}

%
\begin{table}[t]
\centering
{\tabcolsep = 2pt
\small
\begin{tabular}{clc}\toprule
Rank   & Chord symbols & Proportion\\
\midrule
1 to 10 & {\sf C,F,G,Am,Em,Dm{\footnotesize 7},G{\footnotesize 7},Dm,E{\footnotesize 7},Em{\footnotesize 7}} & $79.75\%$\\
11 to 20 & {\sf B$\flat$,Am{\footnotesize 7},F{\footnotesize M7},Fm,A$\flat$,D,C{\footnotesize 7},D{\footnotesize 7},A,A{\footnotesize 7}} & $90.92\%$\\
21 to 50 & {\sf E$\flat$,E,C{\footnotesize M7},G{\footnotesize sus4},Bm{\footnotesize 7{$-$}5},Gm,C{\footnotesize sus4},Gm{\footnotesize 7}} &\\
       & {\sf D$\flat$,Cm,F{\footnotesize 7},F$\sharp$,F$\sharp${\footnotesize dim},A$\flat${\footnotesize dim},F$\sharp${\footnotesize m7$-$5}}&\\
       & {\sf Bm,Fm{\footnotesize 7},B,C$^{\sf add9}$,F$\sharp$m,G{\footnotesize 7sus4},B$\flat$m}&\\
       & {\sf B{\footnotesize 7},F$^{\sf add9}$,Bm{\footnotesize 7},Cm{\footnotesize 7},B{\footnotesize dim},C{\footnotesize 6},B$\flat${\footnotesize 7},G{\footnotesize 6}}& $98.20\%$\\
\bottomrule
\end{tabular}
}
\caption{Most frequent chord symbols in the J-pop data.}
\label{tab:JPopData}
\vspace{-2mm}
\end{table}
Two datasets of chord sequences of popular music pieces were used for comparative evaluation of the models.
The {\it J-pop data} consisted of chord sequences of 3500 J-pop songs that were obtained from a public web page \footnote{J-Total Music: \url{http://music.j-total.net} (Researchers who wish to have access to the J-Pop data should contact the authors.)}.
All songs had key information in the major mode and were transposed to C major key.
Each sequence in this dataset corresponds to a whole piece and the average length was $119.5$ (chords).
The computation time required for learning and inferring PCFG models using such long sequences is practically too large and only Markov models and HMMs were tested using this data.
The other dataset was constructed by extracting chord sequences from the {\it Billboard dataset} \cite{Burgoyne2011}, which is sampled from the Billboard Hot 100 from 1958 to 1991 \footnote{The dataset was downloaded from the McGill Billboard Project webpage: \url{http://ddmal.music.mcgill.ca/research/billboard} (Complete Annotation section).}.
Using the musical structure annotation, 1702 chord sequences with a length larger than 7 were obtained by segmenting 468 popular pieces in units of sections such as verse and chorus.
Using the key information assigned to each section, the tonic of each sequence was transposed to C.
Because the Billboard data has the information about the tonic but not the mode, the obtained sequences included C minor sequences as well as more frequent C major sequences.
The average length of chord sequences was $13.0$, and we tested all types of models using this dataset.

For evaluation, we separate each of the datasets into training data and test data.
For the J-pop data 500 sequences and for the Billboard data one tenth of all sequences (i.e.\ 171 sequences) were randomly chosen as the test data.
The rest of the sequences (3000 sequences for the J-pop data and 1531 sequences for the Billboard data) were used as the training data.
To examine the generalisation ability of the models, randomly chosen subsets of the training data were prepared.
For the J-pop data we prepared training datasets with 30, 300, and 3000 sequences and for the Billboard data 30, 300, and 1531 sequences.

\begin{table}[t]
\centering
{\tabcolsep = 2pt
\small
\begin{tabular}{clc}\toprule
Rank   & Chord symbols & Proportion\\
\midrule
1 to 10 & {\sf C,F,G,Am,B$\flat$,Cm,G{\footnotesize 7},Dm,Dm{\footnotesize 7},A$\flat$} & $58.42\%$\\
11 to 20 & {\sf Em{\footnotesize 7},E$\flat$,Am{\footnotesize 7},Em,F/C,C{\footnotesize 7},F{\footnotesize M7},Fm} & \\
       & {\sf Cm{\footnotesize 7},C{\footnotesize 5}} & $69.55\%$\\
21 to 50 & {\sf F{\footnotesize 7},G{\footnotesize sus4},C{\footnotesize M7},D,A,Gm,D$\flat$,Fm{\footnotesize 7},NC} &\\
       & {\sf D{\footnotesize 7},G{\footnotesize 7sus4},C/E,G{\footnotesize 7sus4}$^{\sf add9}$,C{\footnotesize 1},A$\flat${\footnotesize M7}}&\\
       & {\sf Gm{\footnotesize 7},F$\sharp$,C/G,F$^{\sf add9}$,G/C,B$\flat$/C,A{\footnotesize 7} }& \\
       & {\sf B$\flat${\footnotesize 5},F{\footnotesize 9},C$^{\sf add9}$,C{\footnotesize 9},B,E{\footnotesize 7},F{\footnotesize 6},G/B}& $84.89\%$\\
\bottomrule
\end{tabular}
}
\caption{Most frequent chord symbols in the Billboard data.}
\label{tab:SalamData}
\vspace{-2mm}
\end{table}
For music processing, a subset of chord symbols is usually selected and used as inputs/outputs of the models.
For example, the set of 12 major triads and 12 minor triads with enharmonically inequivalent roots is often used.
Because some other chords appear more frequently than some major/minor triads (e.g.\ {\sf G{\footnotesize 7}} or {\sf Dm{\footnotesize 7}} is usually used much more frequently than {\sf F$\sharp$} or {\sf B$\flat$m}), we choose several sets of most frequent chord symbols for each dataset as listed in Tables \ref{tab:JPopData} and \ref{tab:SalamData}.
We used the top 10, 20, and 50 chord symbols and all other chord symbols were treated as `{\sf {\footnotesize Other}}' so that the size of the symbol space $N_\Omega=\#\Omega$ was $10{+}1$, $20{+}1$, and $50{+}1$.
Note that the Billboard data is more contaminated with chords in C minor or other keys (due to temporary modulations) and the proportion of `{\sf {\footnotesize Other}}' is relatively large.

We compared Markov models, HMMs, and PCFG models with different learning schemes described in Sec.~\ref{sec:Model}.
For Markov models, we tested from first-order to third-order models learned with the additive smoothing (with additive constant $0.1$), KN smoothing, and MKN smoothing.
For HMMs, we tested the following state-space sizes $N_\Gamma$:
\begin{equation}
\{1,2,\ldots,9,10,15,20,25,30,40,\ldots,90,100\}.
\end{equation}
For PCFG models, the tested numbers of possible nonterminals $N_\Delta$ were $\{1,2,\ldots,9,10,15,20\}$.
We remark that the upper limit on the order of Markov models and that on the $N_\Gamma$ or $N_\Delta$ for HMMs or PCFG models were chosen so that the computational cost is near the practical limit.
Each HMM and PCFG model was learned based on the EM and GS algorithms with random initial parameters and we tested for each case with several random number seeds to examine the effect of initialisation.
For HMMs (PCFG models), at most 500 (200) iterations were carried out in the EM algorithm until the likelihood converges.
Whether the relative difference of likelihoods was less than $10^{-5}$ was used as the criterion of convergence.
In the GS algorithm, the maximum likelihood parameter set was chosen after 500 (200) samplings and then applied the EM algorithm with a maximum of 50 iterations.
For PCFG models, initial parameters obtained from HMM parameters as in Eqs.~(\ref{eq:PCFGMimickingHMM1})--(\ref{eq:PCFGMimickingHMM4}), which will be called HMM initialisations, were also tested.
We set $\eta=0.01/N_\Delta$ and the value of $\kappa$ determined from data using Eq.~(\ref{eq:ExpectedLength}) was $0.5416$.
For both HMMs and PCFG models, the Dirichlet parameters used for the GS algorithm were all set to $0.1$.

\subsection{Comparison between HMMs and Markov Models}
\label{sec:MarkovVSHMM}

\subsubsection{Comparison of Predictive Power}

\begin{figure*}[t]
\centering
\subfigure[$N_\Omega=10+1$.]
{\includegraphics[clip,width=1.3\columnwidth]{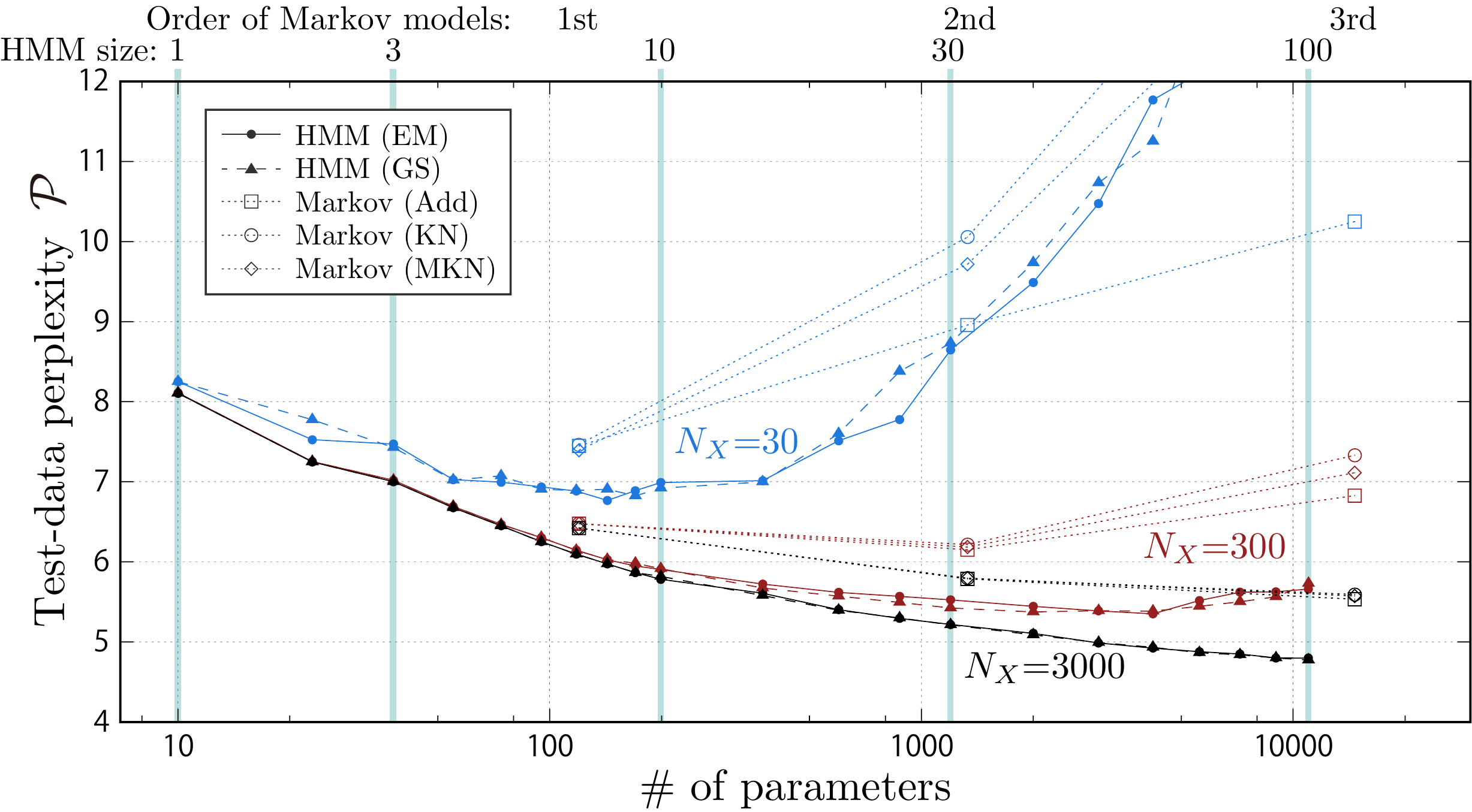}}
\\
\subfigure[$N_\Omega=20+1$.]
{\includegraphics[clip,width=1.3\columnwidth]{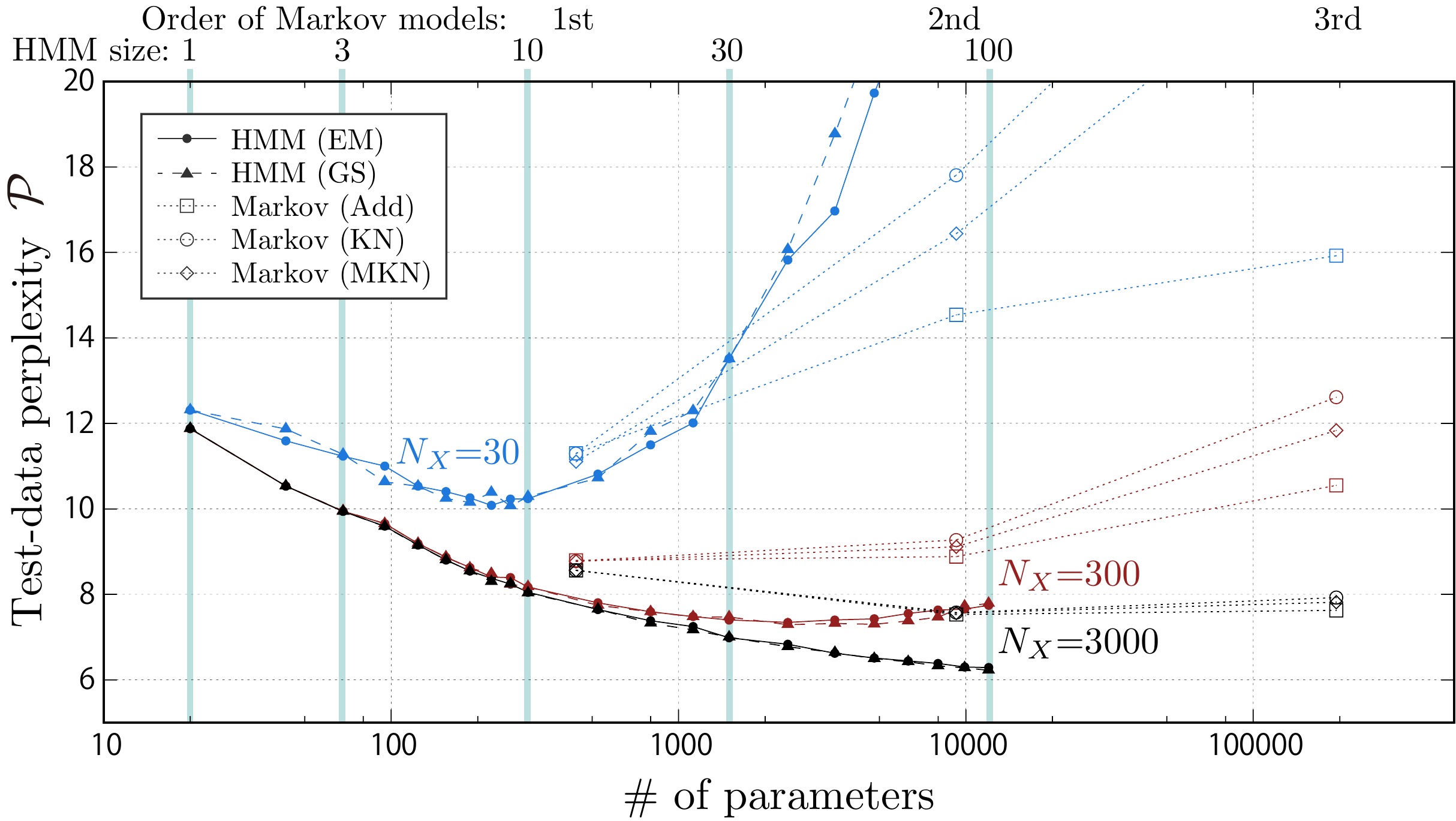}}
\\
\subfigure[$N_\Omega=50+1$.]
{\includegraphics[clip,width=1.3\columnwidth]{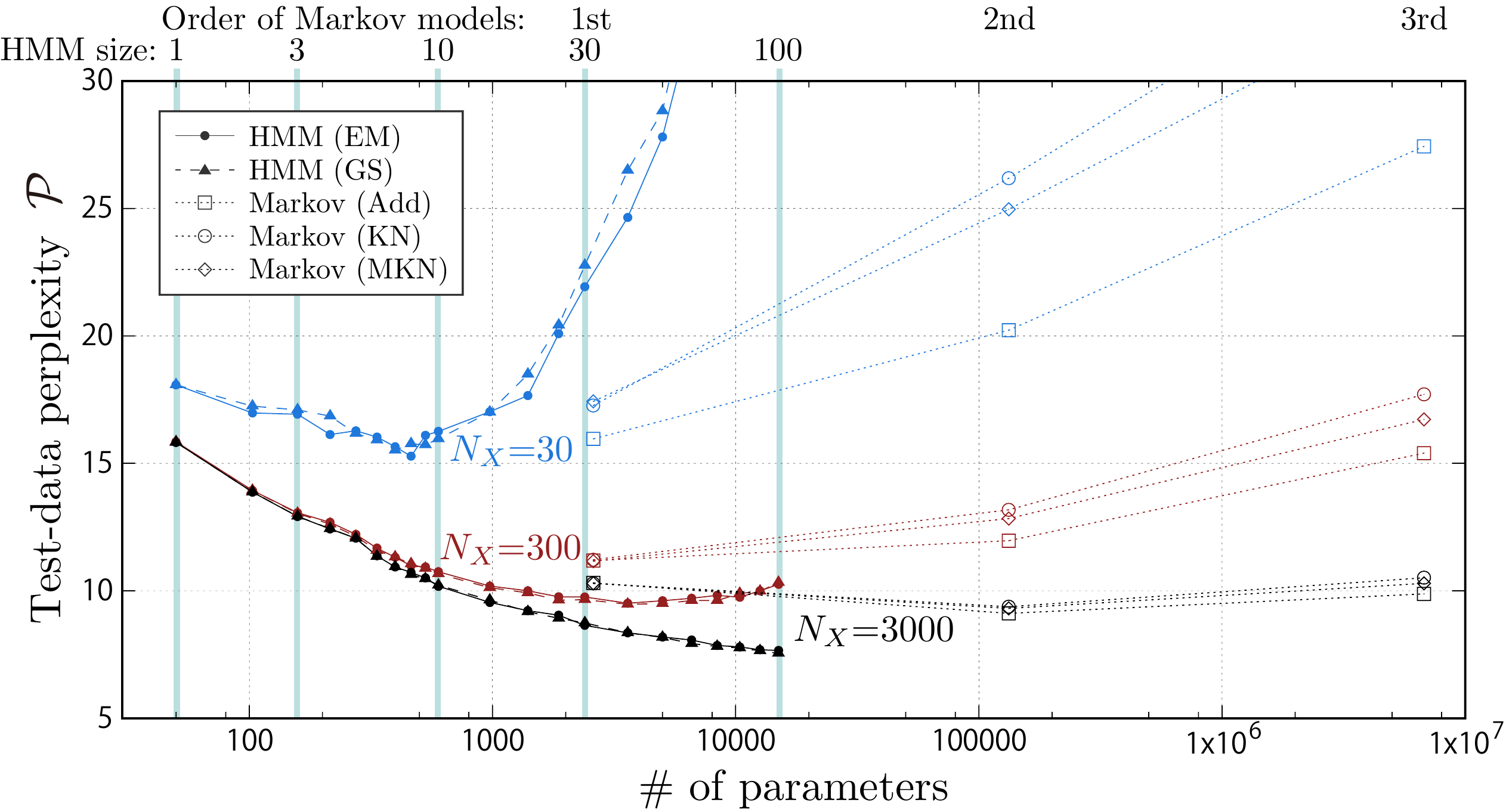}}
\vspace{-3mm}
\caption{Test-data perplexities for the J-pop data. $N_X$ is the number of sequences used for learning and $N_\Omega$ is the number of possible chord symbols. For example, `$N_\Omega=10+1$' means that the most common ten chord symbols were used, plus a symbol `{\sf {\footnotesize Other}}' for all other chords. The horizontal axis indicates the number of parameters given in Table \ref{tab:Parameters} and the corresponding model size (the order of Markov models or the number of states) is indicated on the upper side.}
\label{fig:TestPerp_Jpop}
\vspace{-3mm}
\end{figure*}
\begin{figure*}[t]
\centering
\subfigure[$N_\Omega=10+1$.]
{\includegraphics[clip,width=1.27\columnwidth]{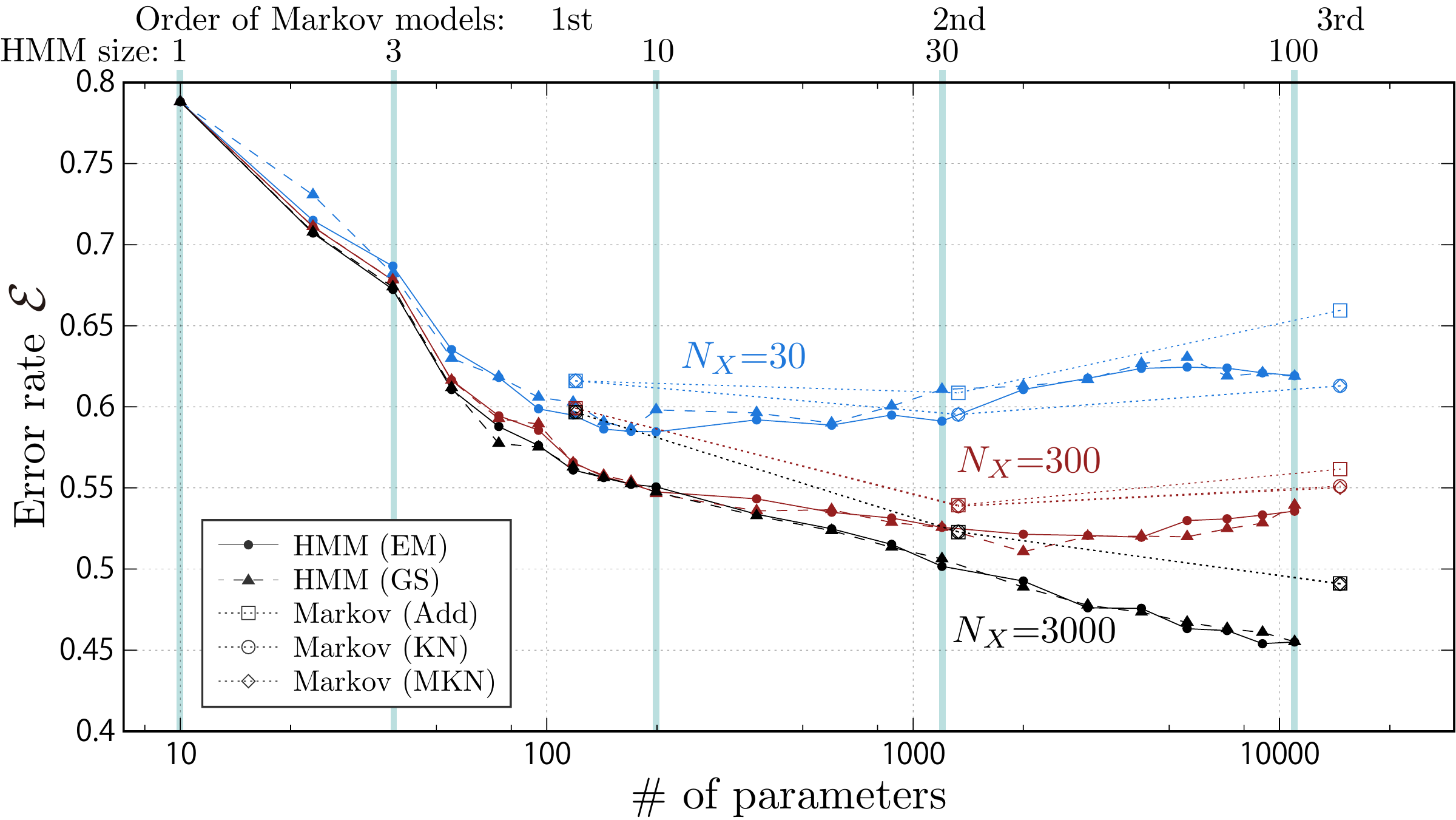}}
\\
\subfigure[$N_\Omega=20+1$.]
{\includegraphics[clip,width=1.27\columnwidth]{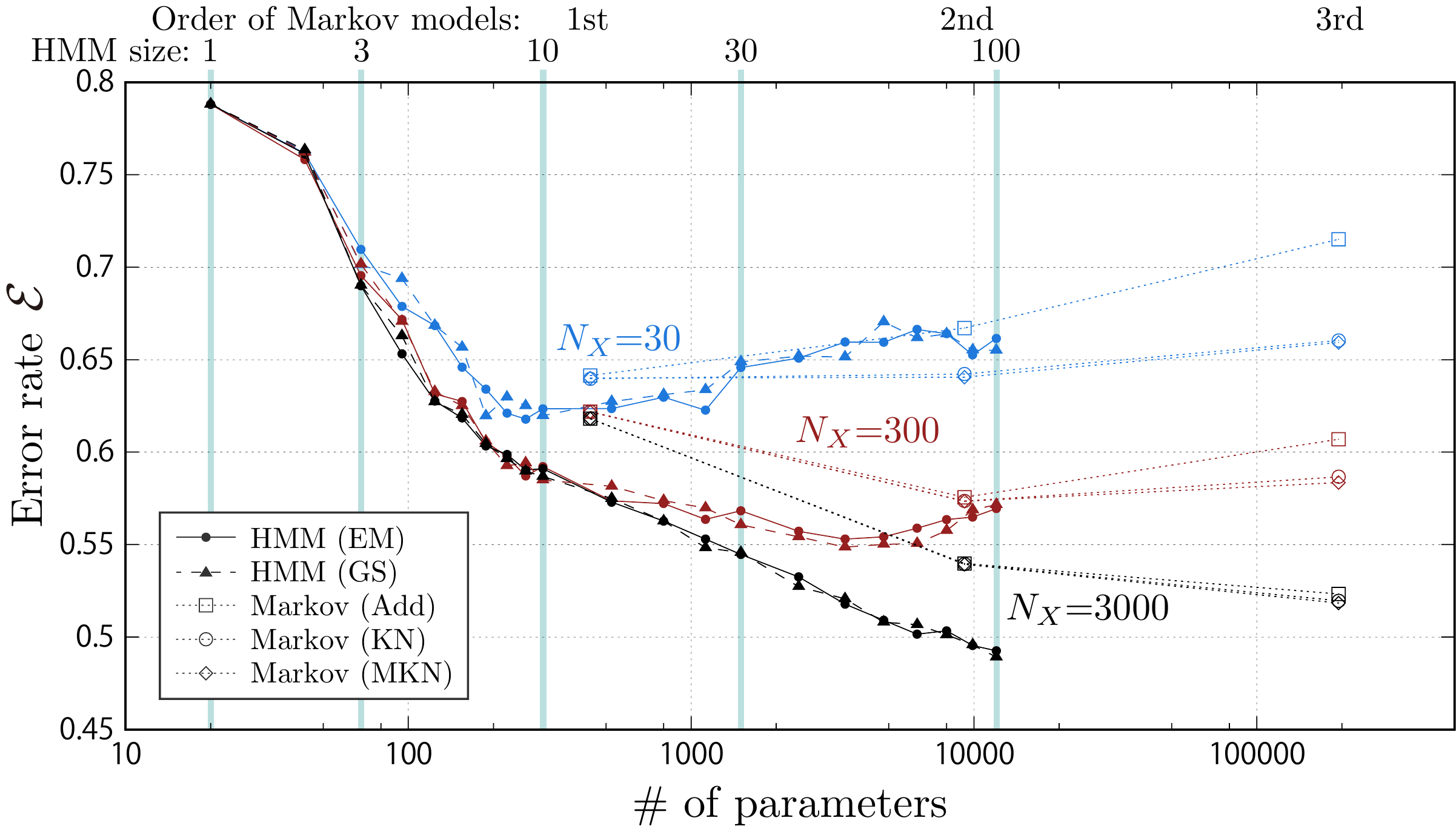}}
\\
\subfigure[$N_\Omega=50+1$.]
{\includegraphics[clip,width=1.27\columnwidth]{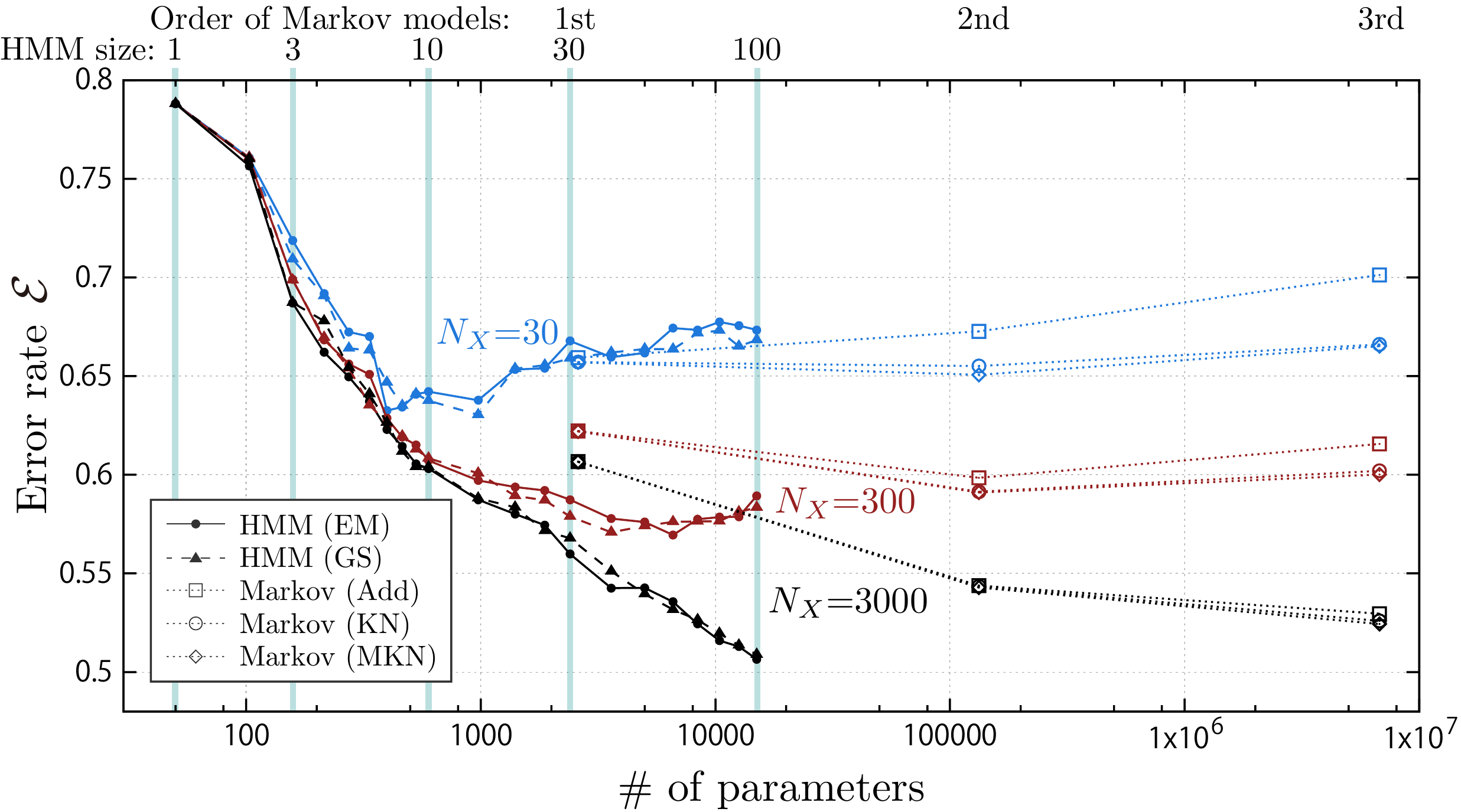}}
\vspace{-2mm}
\caption{Error rates for the J-pop data. See the caption to Fig.~\ref{fig:TestPerp_Jpop} for an explanation of symbols.}
\label{fig:ER_Jpop}
\vspace{10mm}
\end{figure*}
\begin{figure*}[t]
\centering
\subfigure[$N_\Omega=10+1$.]
{\includegraphics[clip,width=1.3\columnwidth]{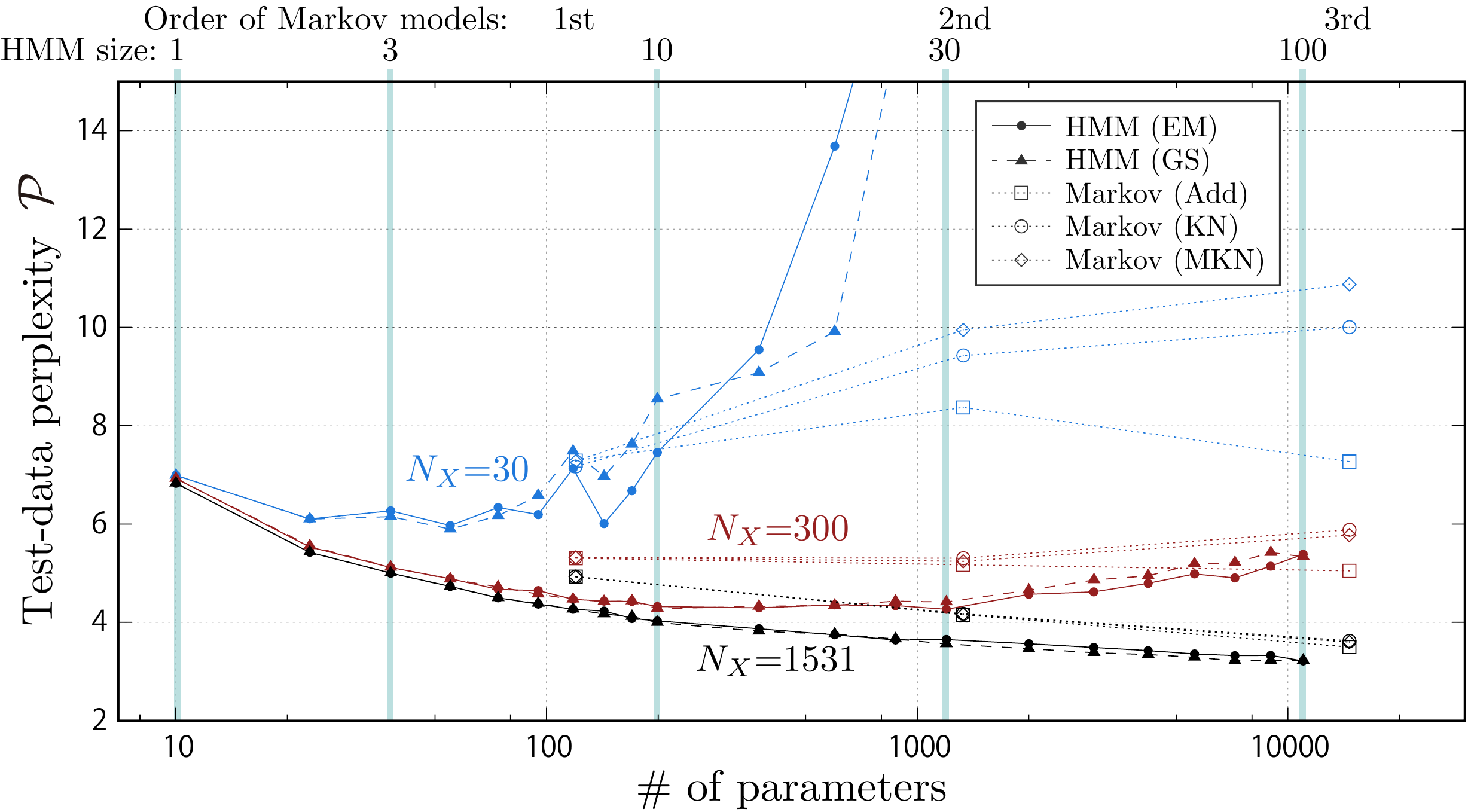}}
\\
\subfigure[$N_\Omega=20+1$.]
{\includegraphics[clip,width=1.3\columnwidth]{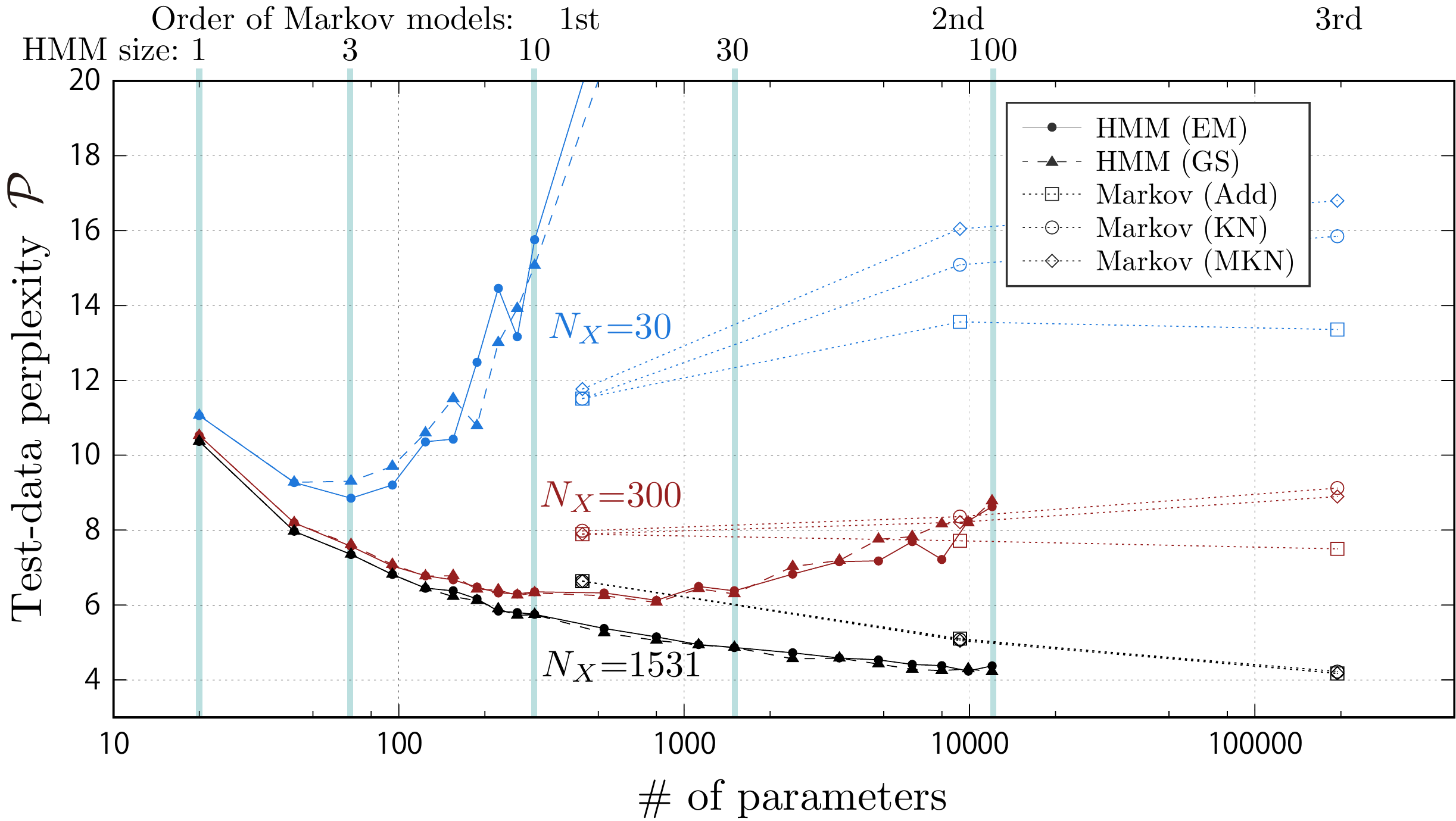}}
\\
\subfigure[$N_\Omega=50+1$.]
{\includegraphics[clip,width=1.3\columnwidth]{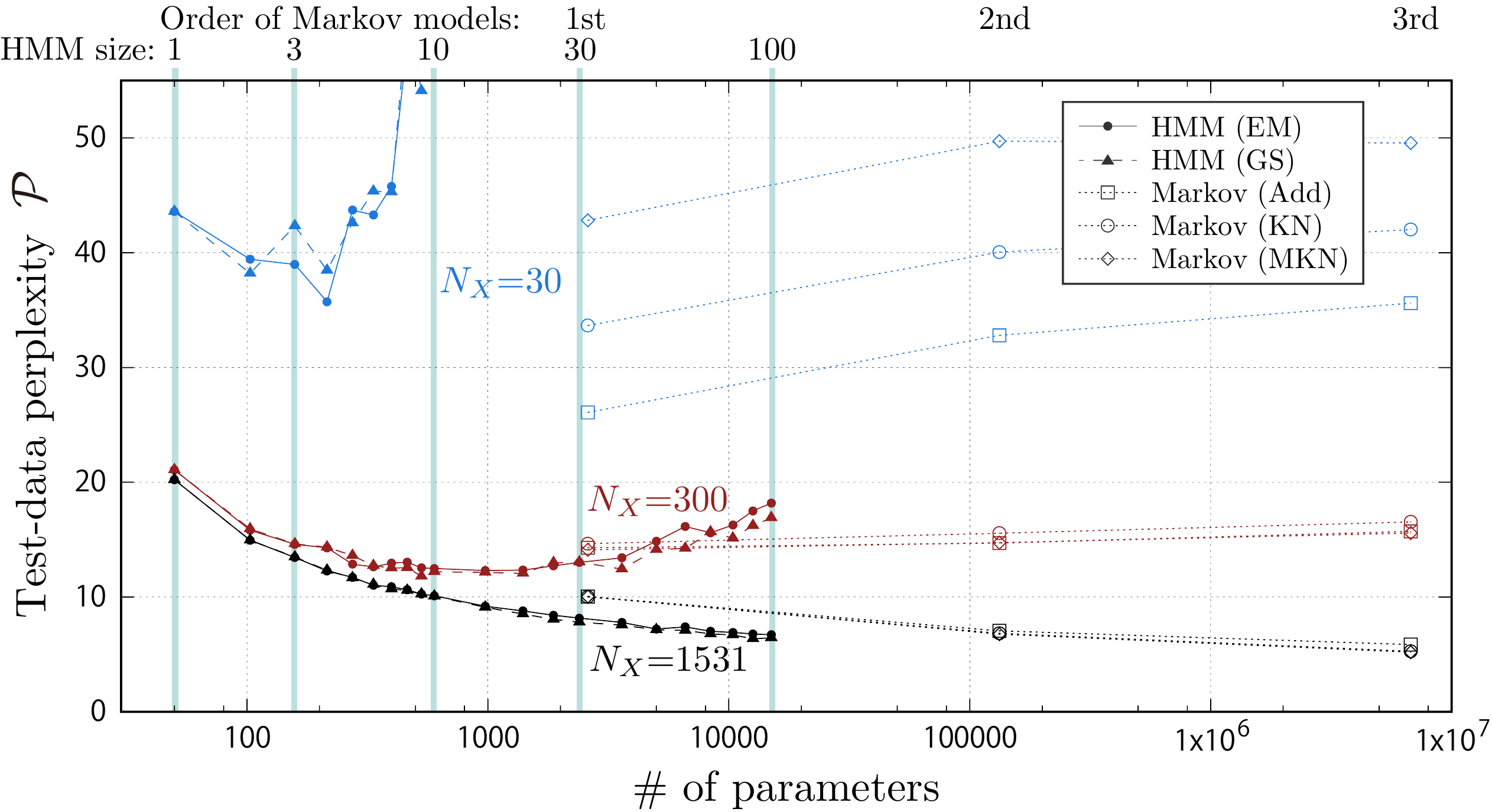}}
\vspace{-3mm}
\caption{Test-data perplexities for the Billboard data. See the caption to Fig.~\ref{fig:TestPerp_Jpop} for an explanation of symbols.}
\label{fig:TestPerp_Salami}
\vspace{10mm}
\end{figure*}
\begin{figure*}[t]
\centering
\subfigure[$N_\Omega=10+1$.]
{\includegraphics[clip,width=1.27\columnwidth]{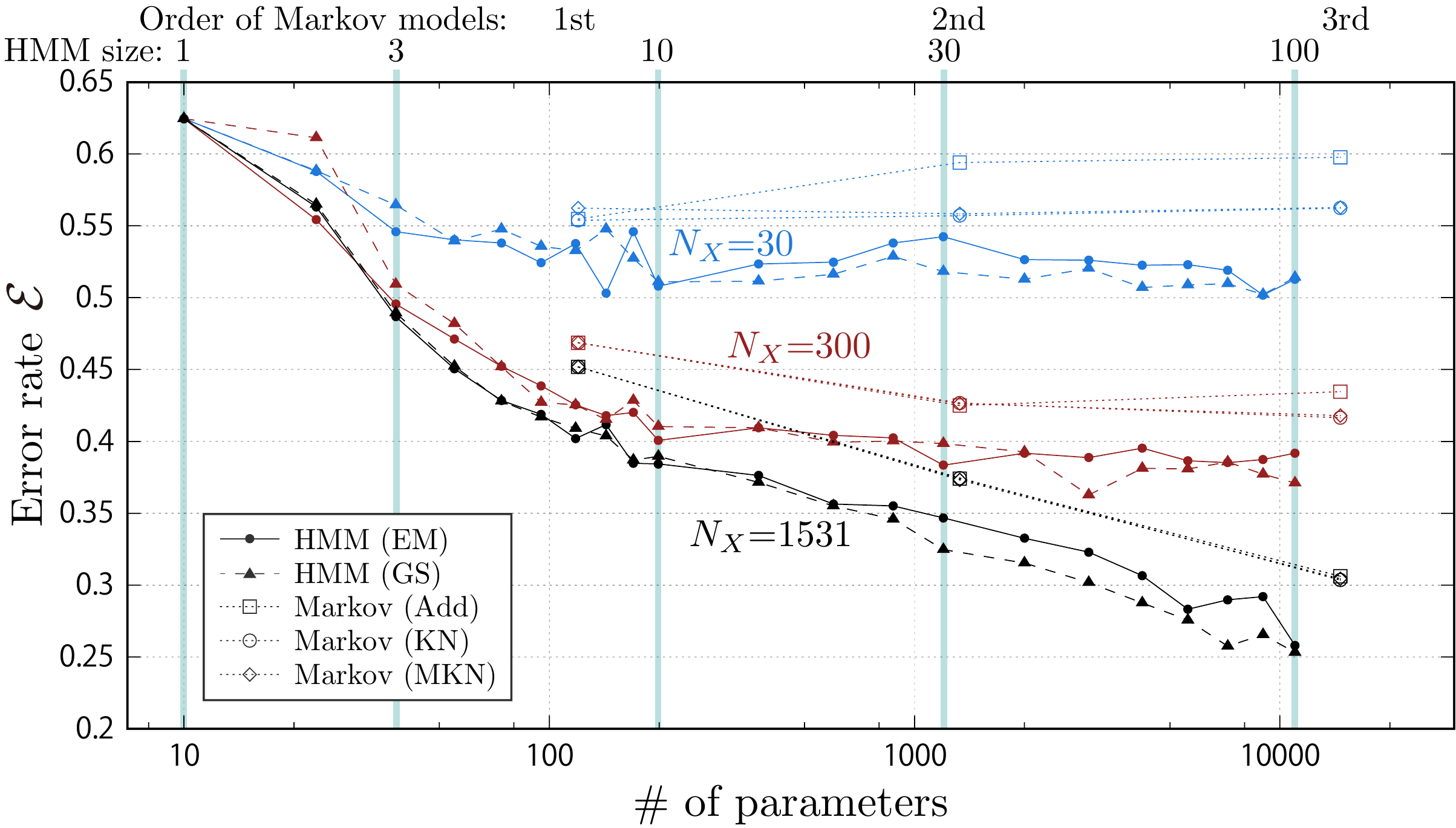}}
\\
\subfigure[$N_\Omega=20+1$.]
{\includegraphics[clip,width=1.27\columnwidth]{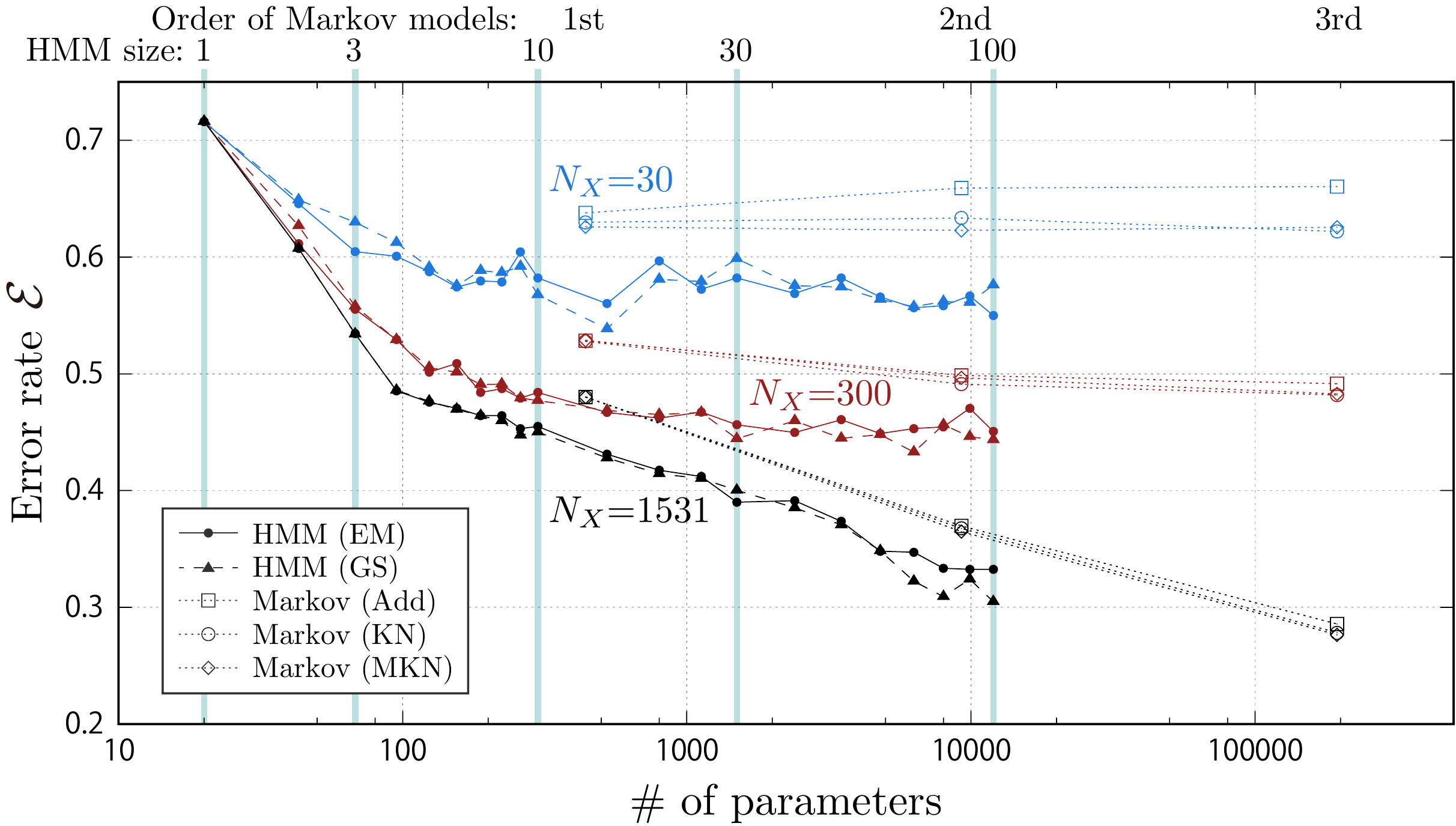}}
\\
\subfigure[$N_\Omega=50+1$.]
{\includegraphics[clip,width=1.27\columnwidth]{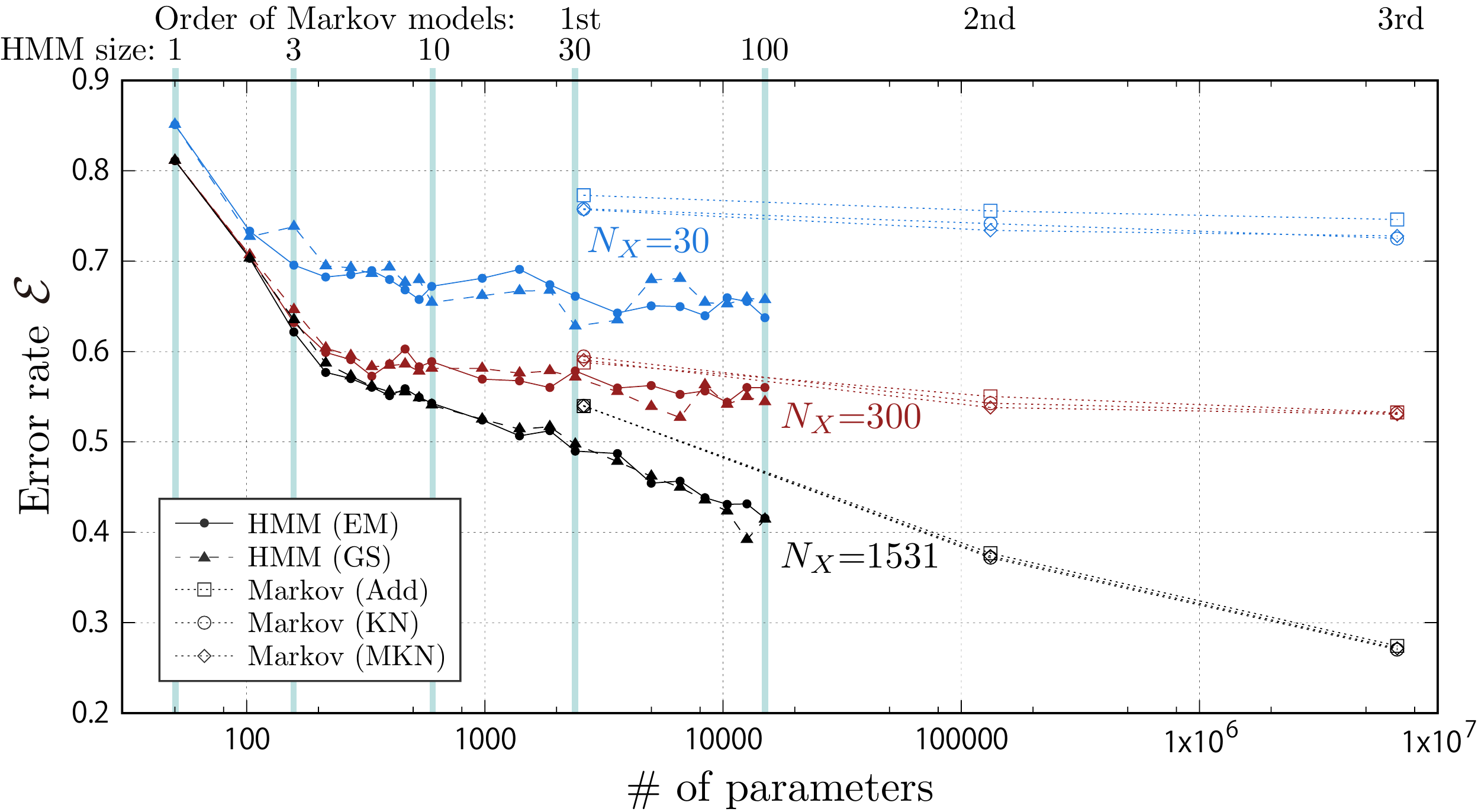}}
\vspace{-2mm}
\caption{Error rates for the Billboard data. See the caption to Fig.~\ref{fig:TestPerp_Jpop} for an explanation of symbols.}
\label{fig:ER_Salami}
\vspace{10mm}
\end{figure*}
Let us first compare the performances of HMMs and Markov models.
The test-data perplexity ${\cal P}$ and the error rate ${\cal E}$ for the two datasets are given in Figs.~\ref{fig:TestPerp_Jpop}, \ref{fig:ER_Jpop}, \ref{fig:TestPerp_Salami}, and \ref{fig:ER_Salami}, 
where the number of sequences in the training data denoted by $N_X$ was 30, 300, and 3000/1531 sequences for the J-pop/Billboard data.
The number of tested random number seeds was 30 when the number of latent states $N_\Gamma\leq 20$ and 10 otherwise.
The displayed perplexity or error rate indicates the best one among the tested random number seeds.

For the J-pop data (Figs.~\ref{fig:TestPerp_Jpop} and \ref{fig:ER_Jpop}), we see that for all $N_X$ and for all $N_\Omega$ (the number of possible chord symbols), the lowest test-data perplexity by the HMMs was less than the lowest value by the Markov models.
For $N_X=30$ and 300, one can find that the test-data perplexity for the HMMs reaches a minimum for some $N_\Gamma$ between 1 and 100 and it increases for larger $N_\Gamma$, which is the overfitting effect.
For $N_X=3000$, both test-data perplexity and error rate decreased up to $N_\Gamma=100$, implying that they could be further reduced with larger size HMMs.
For comparison between the EM and GS algorithms, there were no significant differences between their test-data perplexities in the regions without overfitting.
Some differences between the error rates can be found, but we do not observe clear tendencies.

\begin{figure}[t]
\centering
\subfigure[Test-data perplexity vs error rate.]
{\includegraphics[clip,width=0.98\columnwidth]{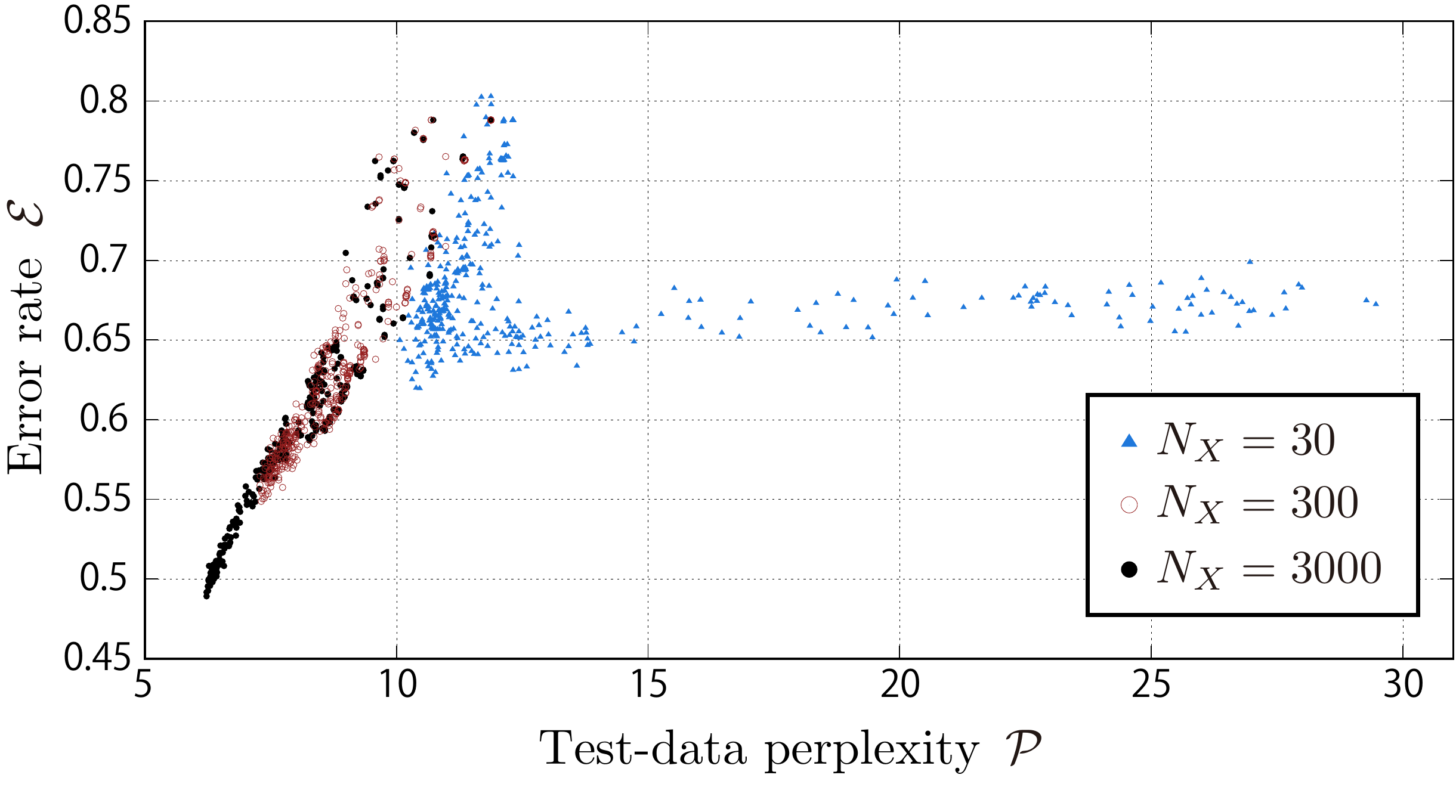}}
\\
\subfigure[RMRR vs error rate.]
{\includegraphics[clip,width=0.98\columnwidth]{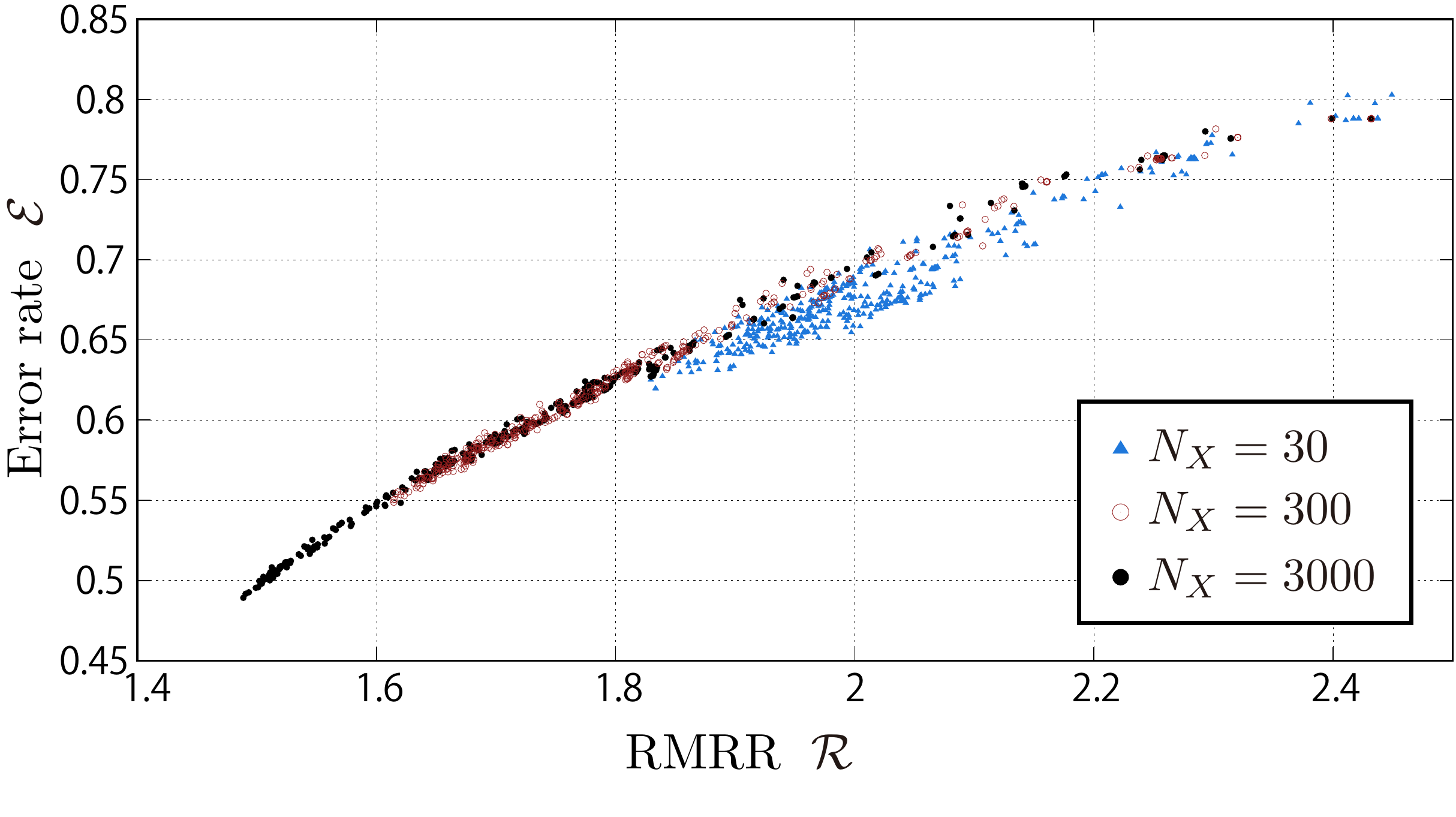}}
\vspace{-3mm}
\caption{Correlations between test-data perplexities, error rates, and RMRRs (reciprocal of the mean reciprocal rank; see Eq.~(\ref{eq:RMRR})) for the J-pop data ($N_\Omega=20+1$, GS). See the caption to Fig.~\ref{fig:TestPerp_Jpop} for an explanation of symbols.}
\label{fig:ERvsRMRR}
\vspace{-3mm}
\end{figure}
For the Billboard data (Figs.~\ref{fig:TestPerp_Salami} and \ref{fig:ER_Salami}), we see a similar tendency of overfitting for $N_X=30$ and $300$ as in the case for the J-pop data.
The lowest test-data perplexity and error rate obtained for HMMs were respectively less than those for Markov models.
However, for $N_X=1531$ and for $N_\Omega=20{+}1$ and $50{+}1$, the results for the third-order Markov models were less than or similar to the best results for the HMMs.
In this case, both test-data perplexity and error rate decreased up to $N_\Gamma=100$ for HMMs and similarly for Markov models up to the third order, so both types of models could yield lower values by further increasing the model sizes.
If the perplexities or error rates for models with similar numbers of parameters are compared, HMMs consistently outperformed Markov models.
For this data, we find a tendency that the results of the GS algorithm were better than those of the EM algorithm for $N_X=1531$ and $N_\Delta\geq50$.

\newpage
The relation between the test-data perplexity and error rate and that between the error rate and RMRR obtained for differently learned HMMs are shown in Fig.~\ref{fig:ERvsRMRR}, for the case of $N_\Omega=20{+}1$ for the J-pop data.
We see that these quantities are highly correlated especially for lower values (corresponding to higher predictive power), indicating that, despite the different definitions, they consistently work as measures of predictive power in the limit when the training-data size and state-space size are large.
In Fig.~\ref{fig:ERvsRMRR}(a), there are two branches of models for $N_X=30$, one of which extends to high test-data perplexities with small changes in the error rate.
This branch corresponds to the overfitting models with $N_\Gamma\gtrsim10$ and is a result of a general tendency that the overfitting has a relatively little effect on the error rate, as seen in Figs.~(\ref{fig:TestPerp_Jpop})--(\ref{fig:ER_Salami}).
Results for other cases and for the Billboard data were similar.

In summary, we confirmed that especially for small training-data sizes and small symbol-space sizes, the predictive power of HMMs exceeds that of Markov models.
For larger training-data size and larger symbol-space size, HMMs might be able to still outperform Markov models if we could increase the model size.
However, increasing the HMM size larger than $N_\Gamma=100$ is practically difficult due to the large computational cost and thus may not be effective for application to music processing.
In comparison between the EM and GS algorithms for learning HMMs, the GS algorithm may perform better especially for a large number of latent states.

\subsubsection{Influence of Initialisation for HMMs}
\label{sec:InfluenceOfInitialisationHMM}

The next problem is how to find the optimal parameter value among those obtained from different initialisations using only the training data.
This is an important issue since, in the practical situation of music processing, one wants to find a model that works best for unseen data.
A natural strategy to estimate the optimal parameter values is to choose the one that minimises the perplexity or error rate evaluated on the training data.

\begin{figure}[t]
\centering
\subfigure[$N_X=30$ and $3000$.]
{\includegraphics[clip,width=0.98\columnwidth]{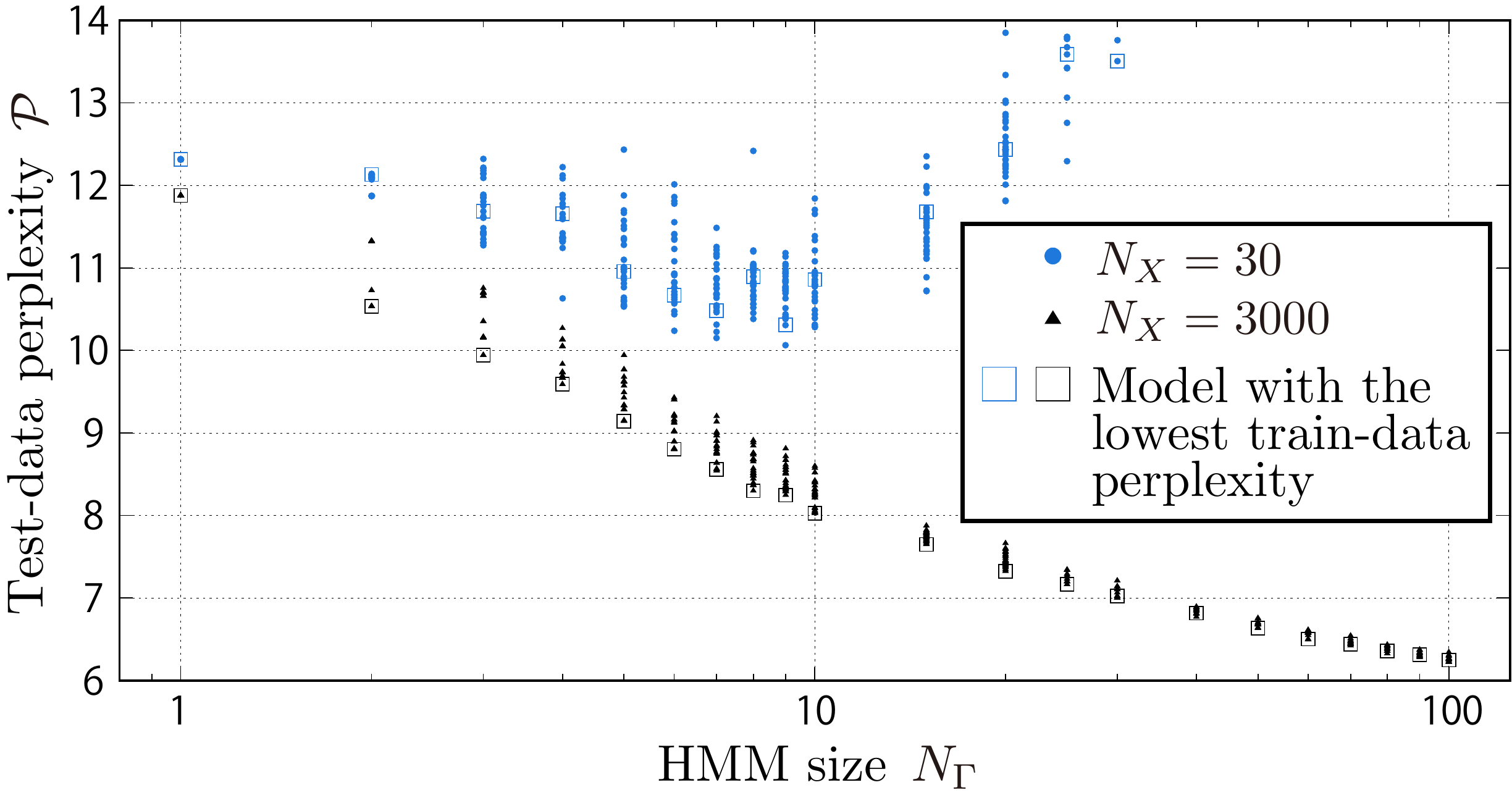}}
\\
\subfigure[$N_X=300$.]
{\includegraphics[clip,width=0.98\columnwidth]{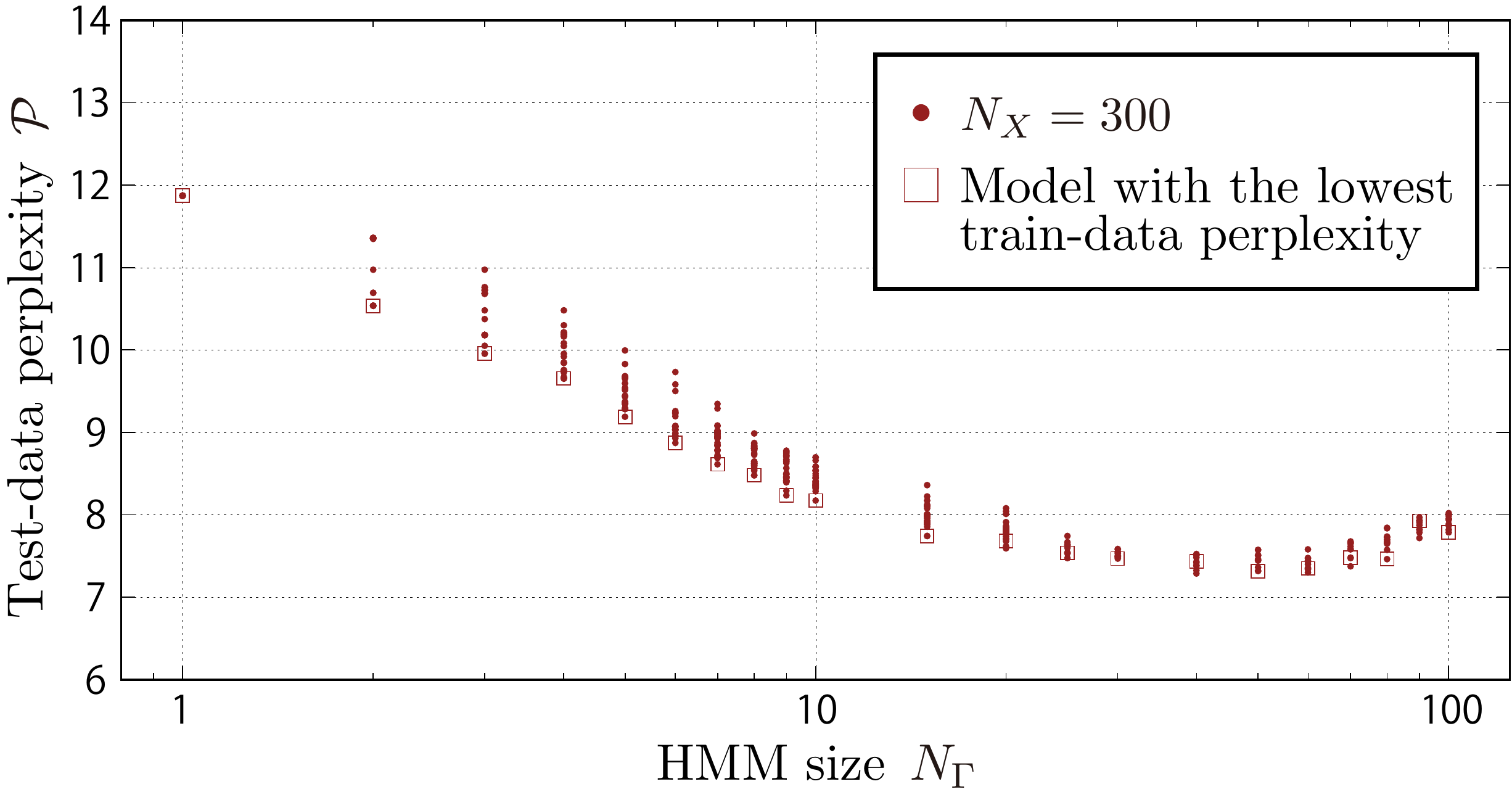}}
\vspace{-2mm}
\caption{Test-data perplexities for the J-pop data with different initialisations ($N_\Omega=20+1$, GS). See the caption to Fig.~\ref{fig:TestPerp_Jpop} for an explanation of symbols.}
\label{fig:Initialisation_Jpop}
\vspace{-3mm}
\end{figure}
Fig.~\ref{fig:Initialisation_Jpop} shows the distributions of test-data perplexities for different initial values for the J-pop data with $N_\Omega=20{+}1$ and $N_X=30$, $300$, and $3000$.
These results were obtained with the GS algorithm and the results with the EM algorithm were similar.
In the figure, each point corresponds to each case of initialisation and the model that had the lowest perplexity evaluated on the training data for each model size is indicated with a square.
For $N_X=30$, the dependencies on the initial values are relatively large and the parameters that are optimal for the training data does not necessarily coincide with those for the test data even for small $N_\Gamma$.
For $N_X=300$, there are still some differences among various initial values but they are smaller than those in the case of $N_X=30$.
There is a clear tendency that for each model size the model optimal in terms of the training-data perplexity is also close to optimal in terms of the test-data perplexity up to about $N_\Gamma=30$, above which the overfitting occurs.
For $N_X=3000$, the differences among various initial values are small for larger model sizes and the correspondence between the optimal values for the training and test data is evident.
Even though it was more obscure, there was a similar tendency for the error rates.

From these results, we conclude that although the best way to choose the optimal parameter set is to directly evaluate HMMs on the ground-truth test data, the optimisation using the training data works reasonably: when the training-data size is small the optimisation can predict wrong choices but with increasing data size the accuracy of optimisation increases and the effect of incorrect choices becomes smaller.
For the data we studied, training data of size $N_X\geq 300$ were able to accurately predict the optimal model for unseen data for model sizes without overfitting.

\subsection{Comparison between PCFG Models and HMMs}
\label{sec:PCFGVSHMM}

Let us now compare the results for HMMs and PCFG models tested on the Billboard data.
For PCFG models, the EM and GS algorithms were run with 30 random number seeds for model sizes $N_\Delta\leq6$ and between 4 and 6 random number seeds for $N_\Delta\geq7$ depending on the computation time required for learning.
To learn PCFG models using the HMM initialisation, the optimal HMM parameter values by the GS algorithm were chosen for each case according to the test-data perplexity and then the EM and GS algorithms were applied.

\begin{figure}[t]
\centering
\subfigure[Test-data perplexity.]
{\includegraphics[clip,width=0.98\columnwidth]{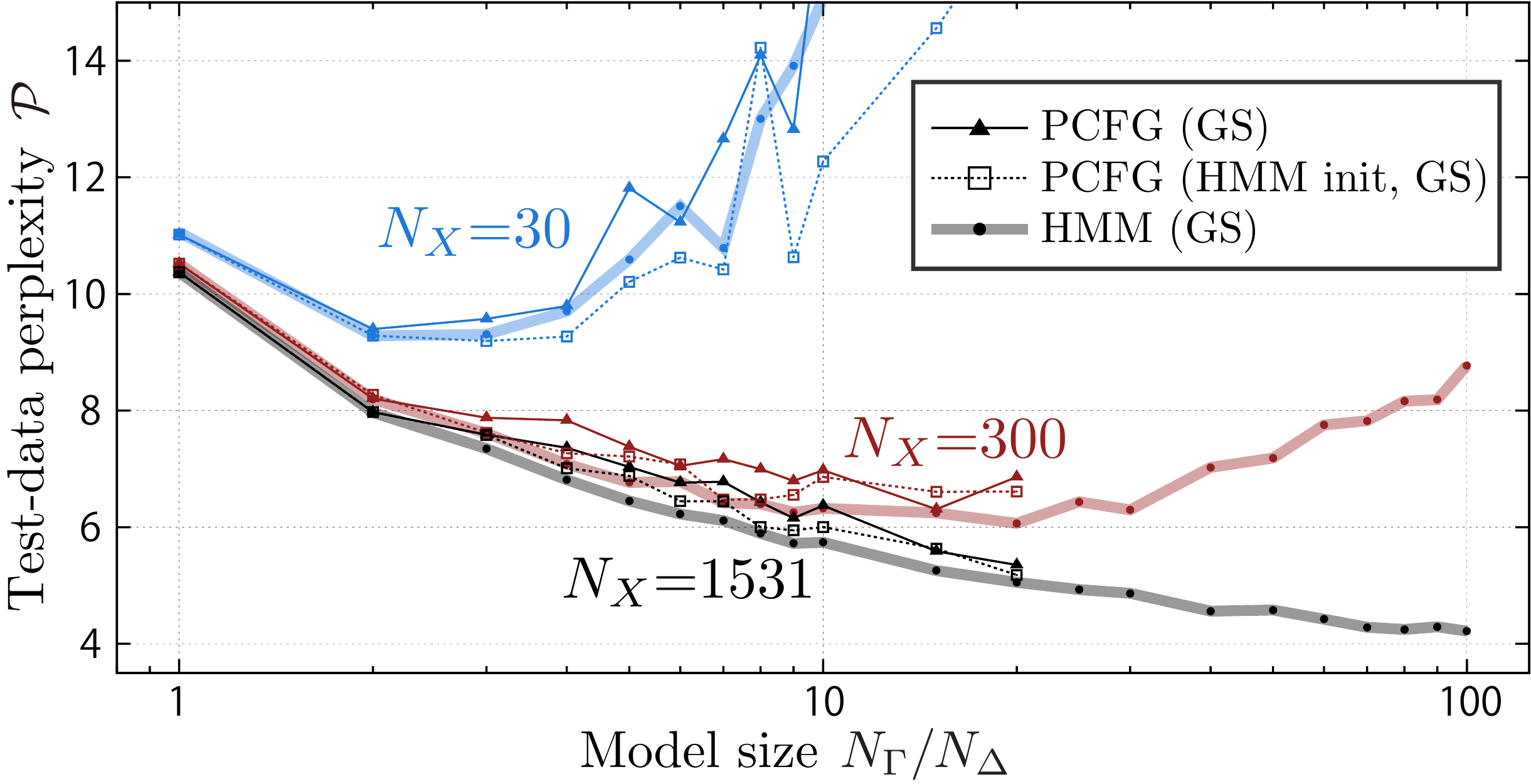}}
\\
\subfigure[Error rate.]
{\includegraphics[clip,width=0.98\columnwidth]{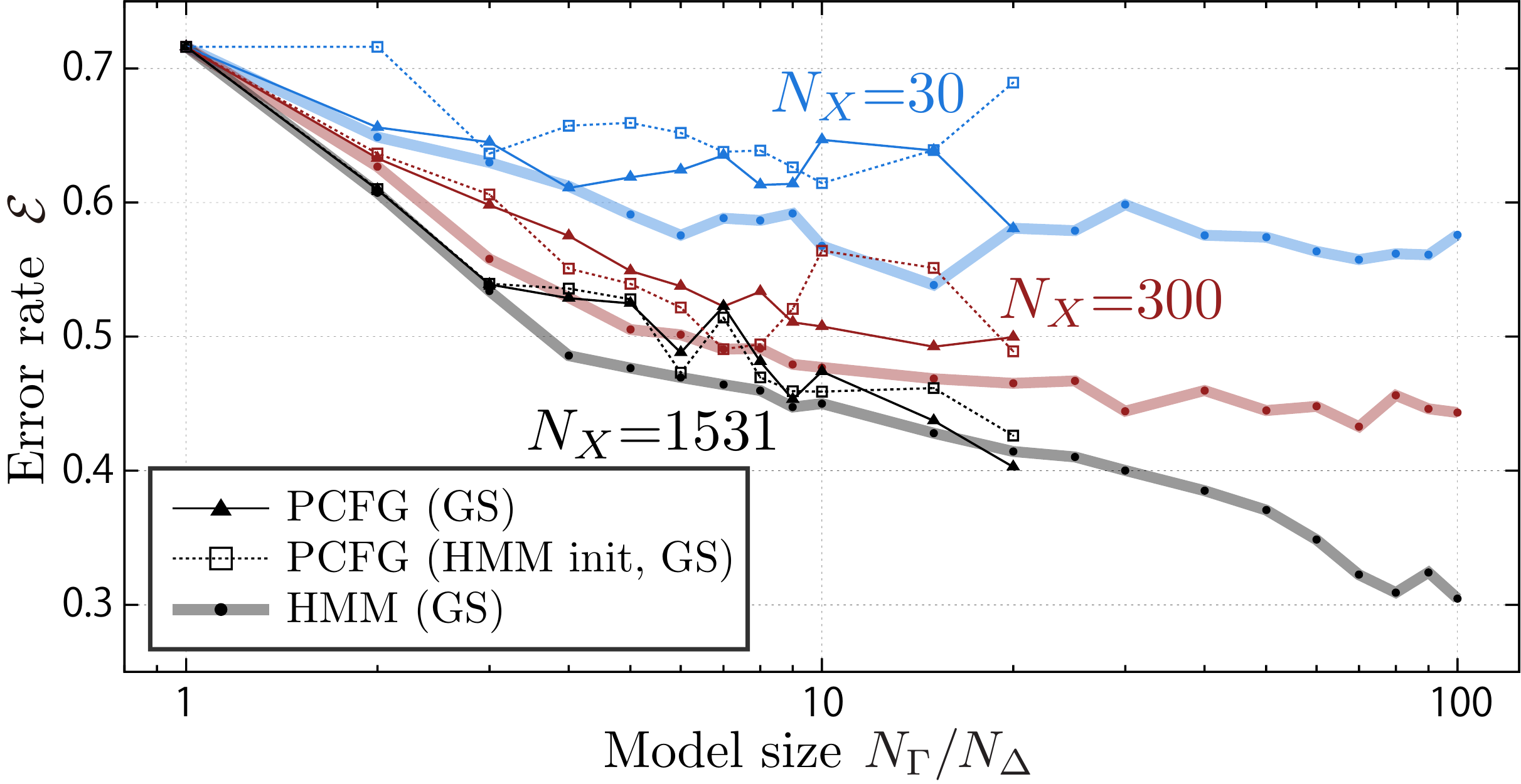}}
\vspace{-2mm}
\caption{Test-data perplexities and error rates for PCFG models trained with the GS algorithm with random and HMM initialisations (Billboard data, $N_\Omega=20+1$). The results for HMMs trained with the GS algorithm are shown as references. See the caption to Fig.~\ref{fig:TestPerp_Jpop} for an explanation of symbols.}
\label{fig:PCFGResults}
\vspace{-3mm}
\end{figure}
In the following, we show and discuss only the results for $N_\Omega=20{+}1$ but the results for other cases were similar.
The test-data perplexities and error rates for the PCFG models trained with the GS algorithm with random and HMM initialisations are shown in Fig.~\ref{fig:PCFGResults}, where for random initialisations the best values are shown.
The results for HMMs with the GS algorithm are also shown as references.

For test-data perplexities, results for PCFG models showed a clear overfitting effect in the case of $N_X=30$.
For $N_X=300$ and $1531$, the PCFG models worked at most equivalently and often worse than the HMMs with $N_\Gamma=N_\Delta$.
There is a clear tendency that the HMM initialisations yielded better results than the random initialisations in the range $4\leq N_\Delta\leq 10$.
These results show the difficulty of parameter optimisation for PCFG models and the effectiveness of the HMM initialisation in some cases.
Notice that for $N_X=30$ and $300$, the lowest test-data perplexity for the PCFG models was less than that of the Markov models.

For error rates, we again find that overall the PCFG models worked at most equivalently and often worse than the HMMs with $N_\Gamma=N_\Delta$.
Even for the case of $N_X=300$ and $1531$, where the overfitting effect was not eminent for the test-data perplexity, we see considerable variations among error rates for different $N_\Delta$\,s.
This is partly due to the limited number of random number seeds and also reflects weak correlation between test-data perplexities and error rates.
The lowest error rate for the PCFG models was less than that of the Markov models for $N_X=30$ and they were even for $N_X=300$.

\begin{figure}[t]
\centering
\subfigure[$N_X=30$ and $1531$.]
{\includegraphics[clip,width=0.98\columnwidth]{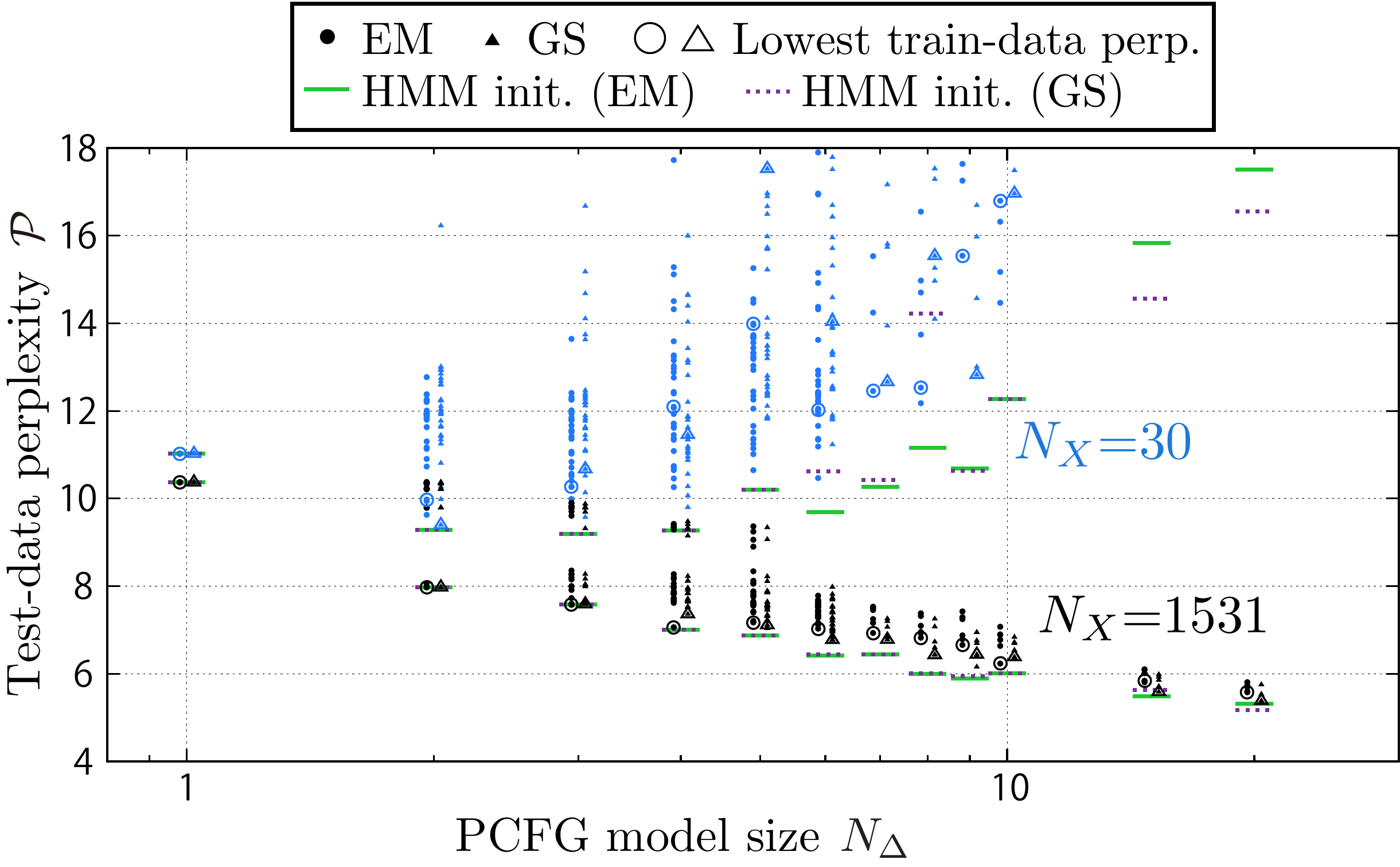}}
\\
\subfigure[$N_X=300$.]
{\includegraphics[clip,width=0.98\columnwidth]{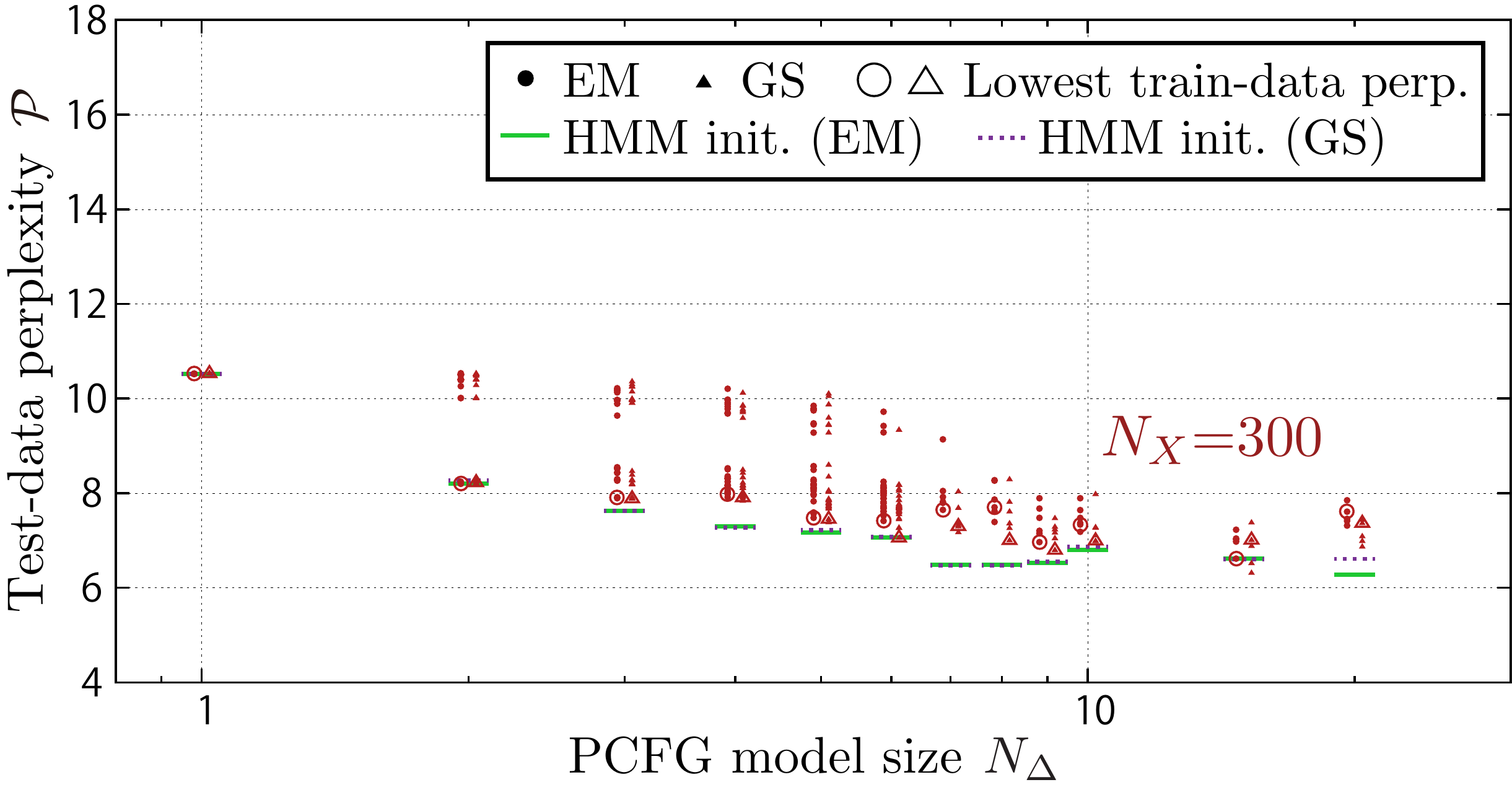}}
\vspace{-2mm}
\caption{Test-data perplexities for PCFG models with different initialisations (Billboard data, $N_\Omega=20+1$). See the caption to Fig.~\ref{fig:TestPerp_Jpop} for an explanation of symbols.}
\label{fig:Initialisation_PCFG}
\vspace{-3mm}
\end{figure}
Let us look more closely at the influence of initialisation and compare the EM and GS algorithms (Fig.~\ref{fig:Initialisation_PCFG}).
For random initialisations, the overall tendency is similar as the case of HMMs: variations in test-data perplexities tend to decrease for larger $N_X$ and, for $N_X=300$ and $1531$, the optimal model according to training-data perplexity also has low test-data perplexity (except for the cases of overfitting).
For $N_X=300$ and $1531$ and $N_\Delta\geq5$, there is a clear tendency that the GS algorithm often found better parameters than the EM algorithm, even though the number of samples were limited and there were exceptions.
On the other hand, for HMM initialisations, differences between the results for the GS and EM algorithms were often negligible.
These results again confirm that there are many local optima for the PCFG-model parameters and it is difficult to find good optima by random initialisation even with the GS algorithm.

\subsection{Structure of Self-Emergent Grammar}
\label{sec:StructureOfSelfEmergentGrammar}

\subsubsection{Information-Theoretical Analysis}
\label{sec:InfTheorAnalysis}
\begin{figure}[t]
\centering
\subfigure[${\cal P}_\phi$ and ${\cal P}_\phi+{\cal P}_\pi$.]
{\includegraphics[clip,width=0.95\columnwidth]{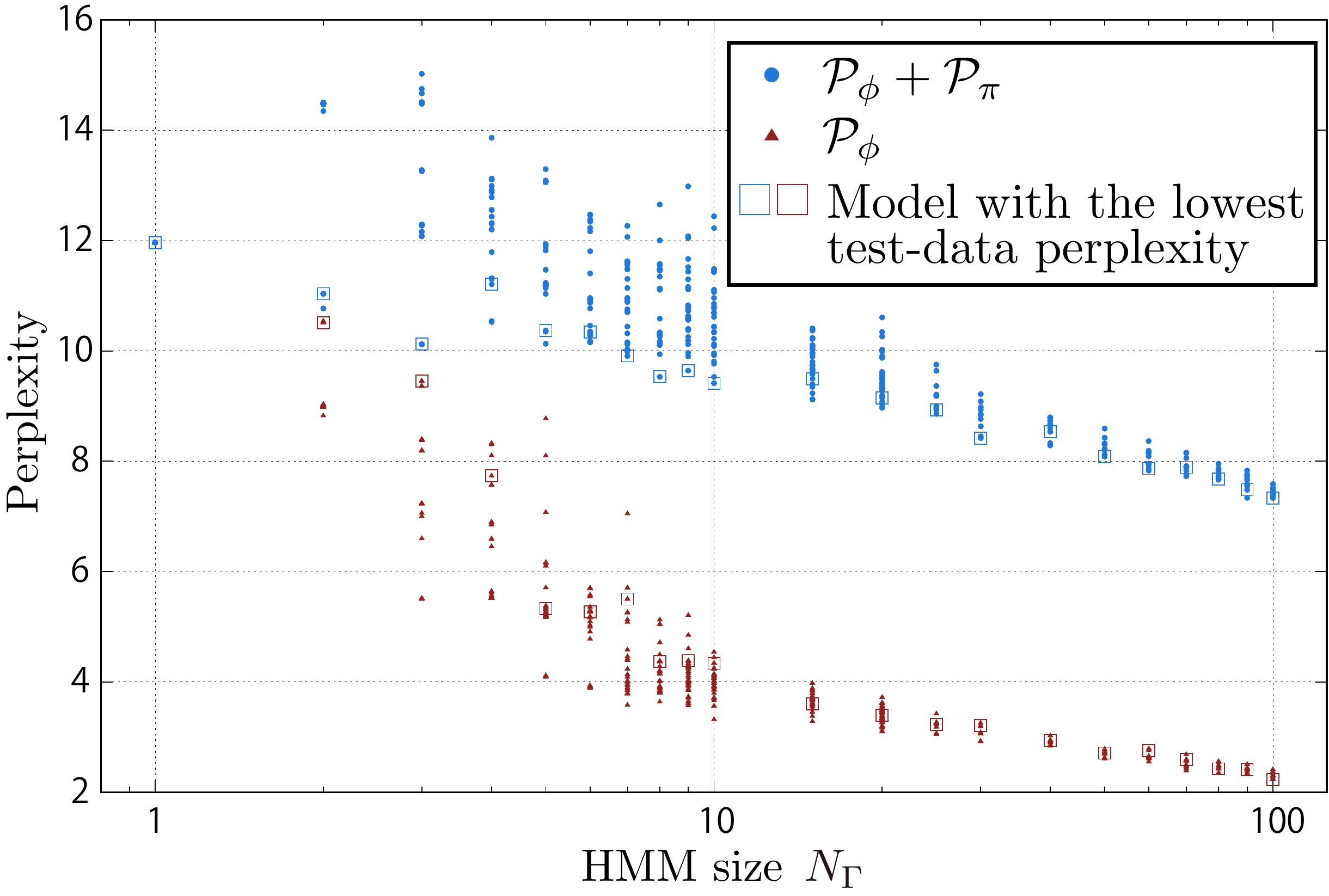}}
\\
\subfigure[${\cal P}_\pi$.]
{\includegraphics[clip,width=0.95\columnwidth]{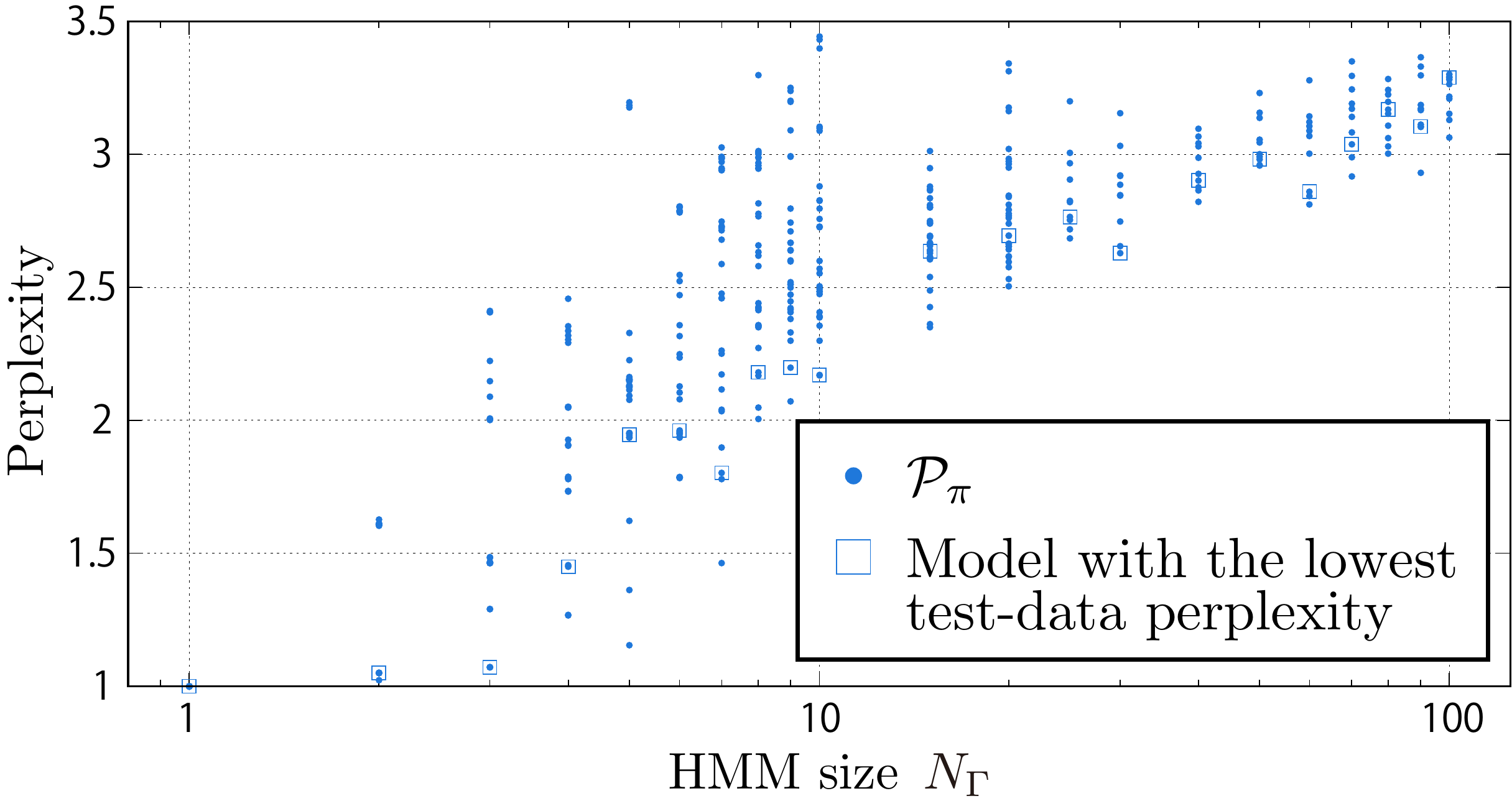}}
\\
\subfigure[${\cal P}_\Gamma$ and ${\cal V}$.]
{\includegraphics[clip,width=0.95\columnwidth]{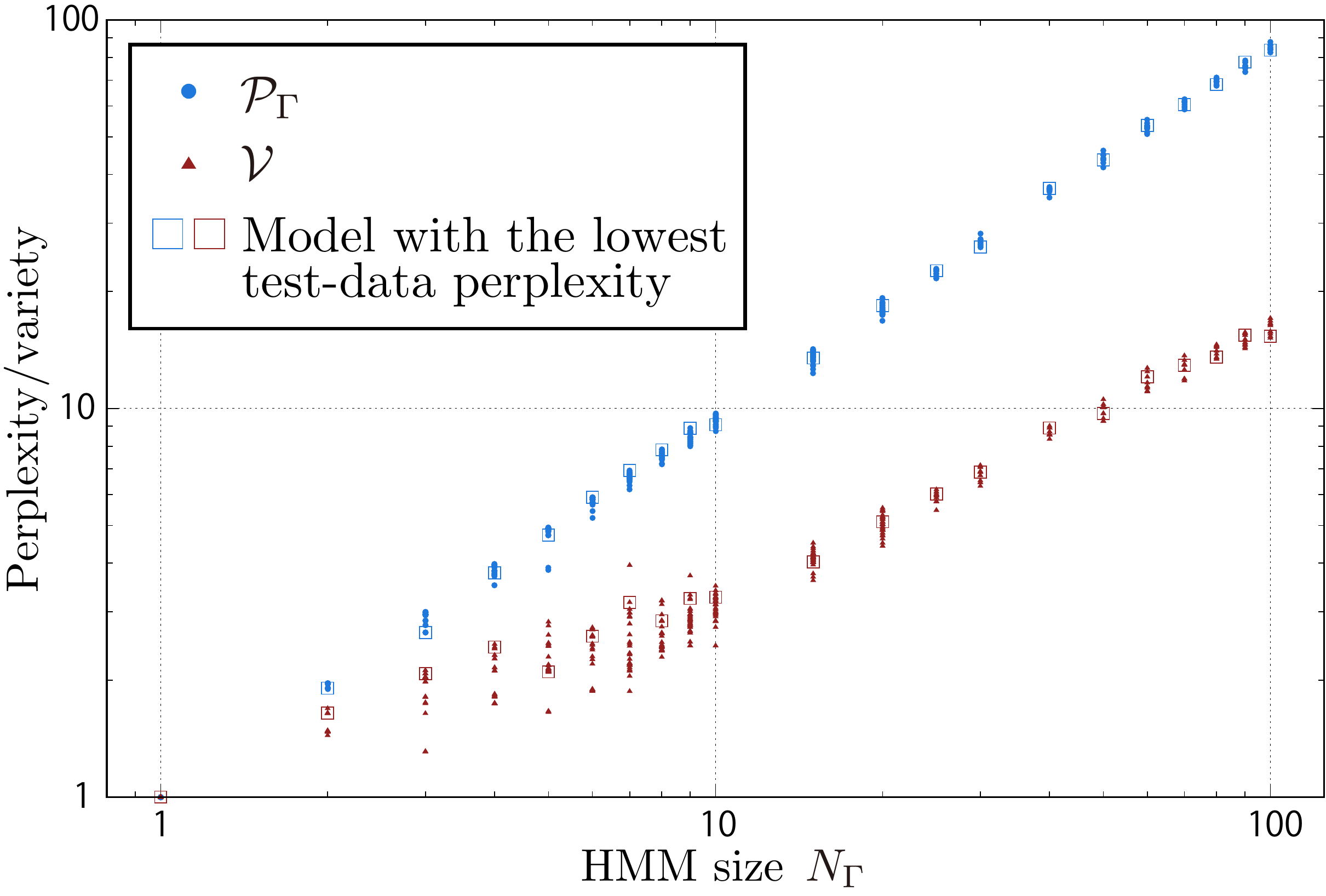}}
\vspace{-2mm}
\caption{Information-theoretical analyses of HMMs trained with the J-pop data ($N_\Omega{=}20{+}1$, GS, $N_X{=}3000$). See Sec.~\ref{sec:HMMAnalysis} for the definition of the information measures ${\cal P}_\phi$, ${\cal P}_\pi$, ${\cal P}_\Gamma$, and ${\cal V}$.}
\label{fig:JPop_HMM_Analysis}
\vspace{-3mm}
\end{figure}
Let us now examine the learned HMMs using the techniques developed in Sec.~\ref{sec:HMMAnalysis}.
The quantities ${\cal P}_\phi$ (average perplexity of output probabilities), ${\cal P}_\pi$ (average perplexity of transition probabilities), ${\cal P}_\phi+{\cal P}_\pi$, ${\cal P}_\Gamma$ (perplexity of the stationary distribution), and ${\cal V}$ (average state association variety) calculated for HMMs learned with the J-pop data with $N_\Omega=20+1$ and $N_X=3000$ are shown in Fig.~\ref{fig:JPop_HMM_Analysis}.
First, the perplexity of the stationary distribution ${\cal P}_\Gamma$ increases roughly proportionally to the model size.
This means that all latent states are exploited almost uniformly to capture the sequential dependence.

Second, the average perplexity of output probabilities ${\cal P}_\phi$ tends to decrease with increasing model sizes, which confirms our expectation that the larger-size models have sparser output probabilities to increase the predictive power.
At the same time, ${\cal P}_\phi$ remains larger than 2 even for $N_\Gamma=100$.
As the HMMs could have ${\cal P}_\phi=1$ for $N_\Gamma\geq21$, this is an evidence that the HMMs automatically captured syntactic similarities among different chord symbols, which was preferred in order to reduce the perplexity.
On the other hand, the increase of the average state assign variety ${\cal V}$ indicates that the same chord symbols are represented by multiple latent states depending on their context, which has a similar effect as symbol refinement for natural language models \cite{Johnson1998,Matsuzaki2005}.

Lastly, the perplexity of transition probabilities ${\cal P}_\pi$ only slowly increases for larger model sizes compared to the increase of ${\cal P}_\Gamma$.
This is another evidence that the larger-size models tend to capture longer-range sequential dependence by using different states to describe different contexts.
We also see that the sum ${\cal P}_\phi+{\cal P}_\pi$, which serves as an approximation of the training-data perplexity, decreases for larger model sizes.
Among models with different initialisations, there is a tendency that the one that minimises the test-data perplexity has low ${\cal P}_\phi+{\cal P}_\pi$, which indirectly implies that the optimal model is determined by balancing low perplexities of output and transition probabilities.

\subsubsection{Examples of Learned Chord Categories}

\begin{figure}[t]
\begin{center}
\subfigure[HMM ($N_\Omega=50+1$, $N_\Gamma=4$, GS, $N_X=3000$).]
{\includegraphics[clip,width=0.99\columnwidth]{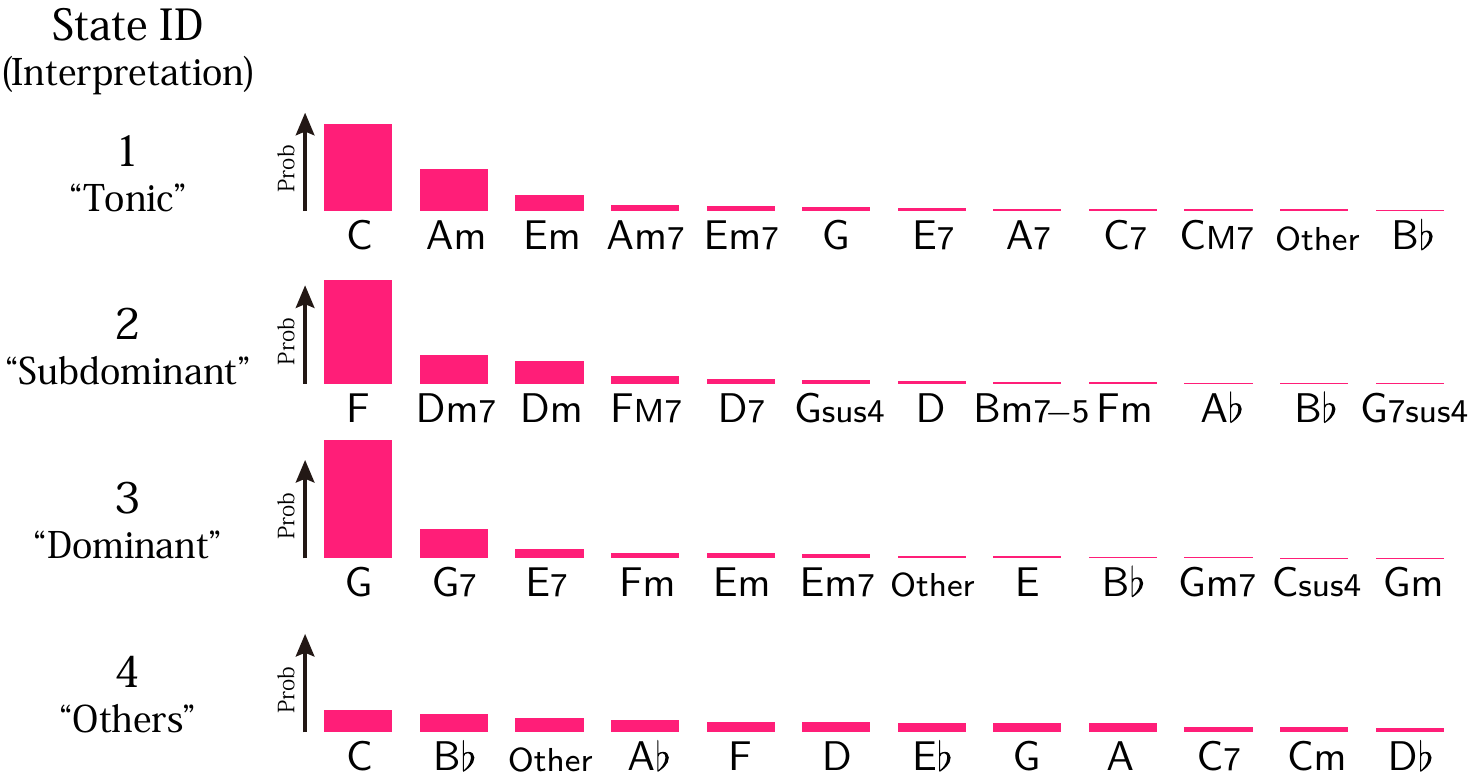}}
\\
\subfigure[PCFG model ($N_\Omega=50+1$, $N_\Delta=5$, GS, $N_X=1531$).]
{\includegraphics[clip,width=0.99\columnwidth]{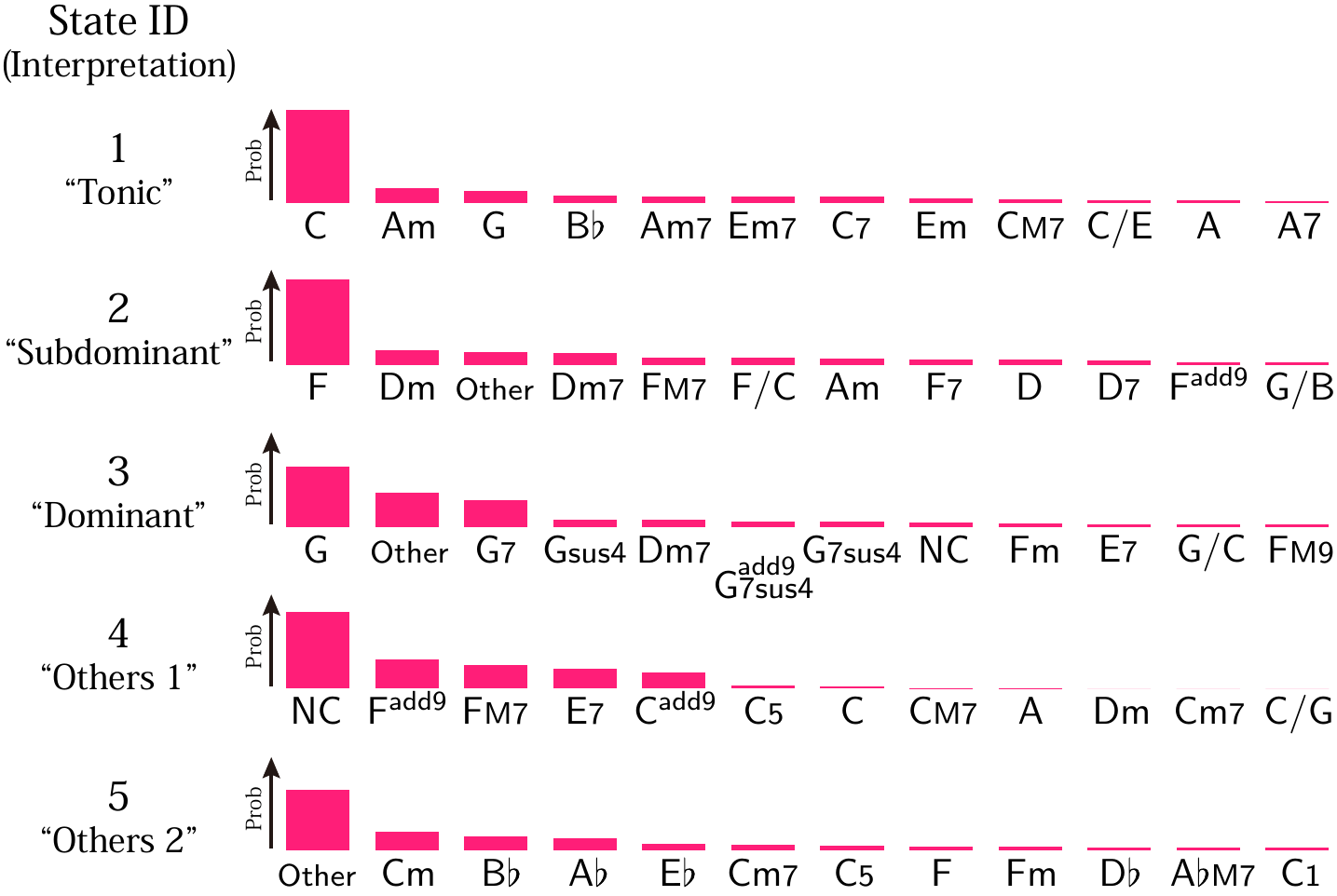}}
\vspace{-5mm}
\end{center}
\caption{Output probabilities of (a) an HMM learned with the J-pop data and (b) a PCFG model learned with the Billboard data. States are manually sorted for clearer illustration and for each state 12 chord symbols with highest output probabilities are shown.}
\label{fig:LearnedCategories}
\end{figure}
An interesting feature of the unsupervised learning of HMMs and PCFG models is that the self-emergent grammars often represent harmonic functions of chords in traditional harmony theories.
Although the output probabilities do not strictly categorise or discriminate chord symbols, we can identify each latent state or nonterminal as a `category' that associates with some chord symbols more strongly than others.
Fig.~\ref{fig:LearnedCategories} shows examples of output probabilities for an HMM trained with the J-pop data and a PCFG model trained with the Billboard data.
The first three states/nonterminals in these examples can be identified as the tonic, subdominant, and dominant functions.
We can see the general tendency that extended chords belong to the same category as a triad with the same root note (e.g.\ {\sf G} and {\sf G{\footnotesize 7}}, and {\sf Dm} and  {\sf Dm{\footnotesize 7}}), and that chords in the parallel relation (e.g.\ {\sf C} and {\sf Am}, and {\sf F} and {\sf Dm}) belong to the same category.
An interesting observation is that {\sf Em} is more strongly associated to the `tonic' category than the `dominant' category in both cases.
Note that these categories are obtained without referring to the constituent pitches of chord symbols.
In both examples, there are one or two categories that contain the symbols {\sf\footnotesize Other} and {\sf NC} as well as relatively rare chords in C major key (e.g.\ {\sf B$\flat$}, {\sf A$\flat$}, {\sf Cm}).

\begin{figure}[t]
\begin{center}
\includegraphics[clip,width=0.95\columnwidth]{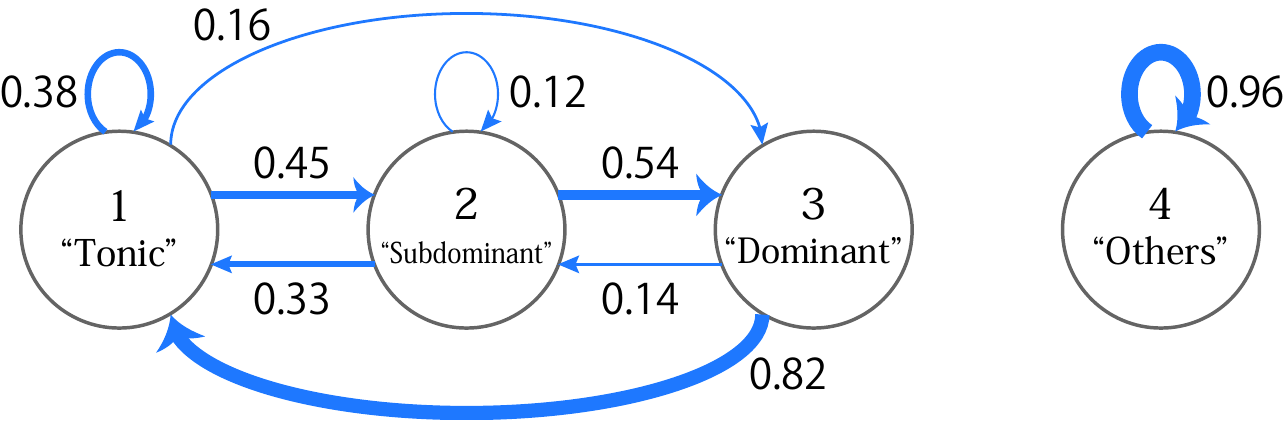}
\vspace{-4mm}
\end{center}
\caption{Learned transition probabilities of the HMM in Fig.~\ref{fig:LearnedCategories}(a). Only probabilities $>0.05$ are indicated.}
\label{fig:LearnedTrProb}
\vspace{-3mm}
\end{figure}
Fig.~\ref{fig:LearnedTrProb} shows the transition probabilities of the 4-state HMM trained with the J-pop data (same as in Fig.~\ref{fig:LearnedCategories}(a)), where only those probabilities larger than $0.05$ are shown for clarity.
This Markov model can be interpreted as a probabilistic representation of the familiar cadence structure, with strong dominant motion (`Dominant' to `Tonic') and rare motion from `Dominant' to `Subdominant'.

Depending on initial parameters, for both HMMs and PCFG models, the learned categories were not always related to traditional harmonic functions as clearly as the above examples.
This can be understood by the existence of multiple local optima in the parameter space as discussed in Sec.~\ref{sec:InfluenceOfInitialisationHMM}.
This may also explain the report \cite{Jacoby2015} that HMMs did not yield chord categories consonant with traditional harmonic functions for some data, since the influence of initialisation was not studied there.
It would be interesting to study further how the traditional harmonic functions can be characterised as a chord categorisation shceme based on information measures and whether alternative forms of chord categorisation are possible.

\subsection{Discussion}
\label{sec:Discussion}

%
\begin{figure*}[t]
\begin{center}
\includegraphics[clip,width=1.5\columnwidth]{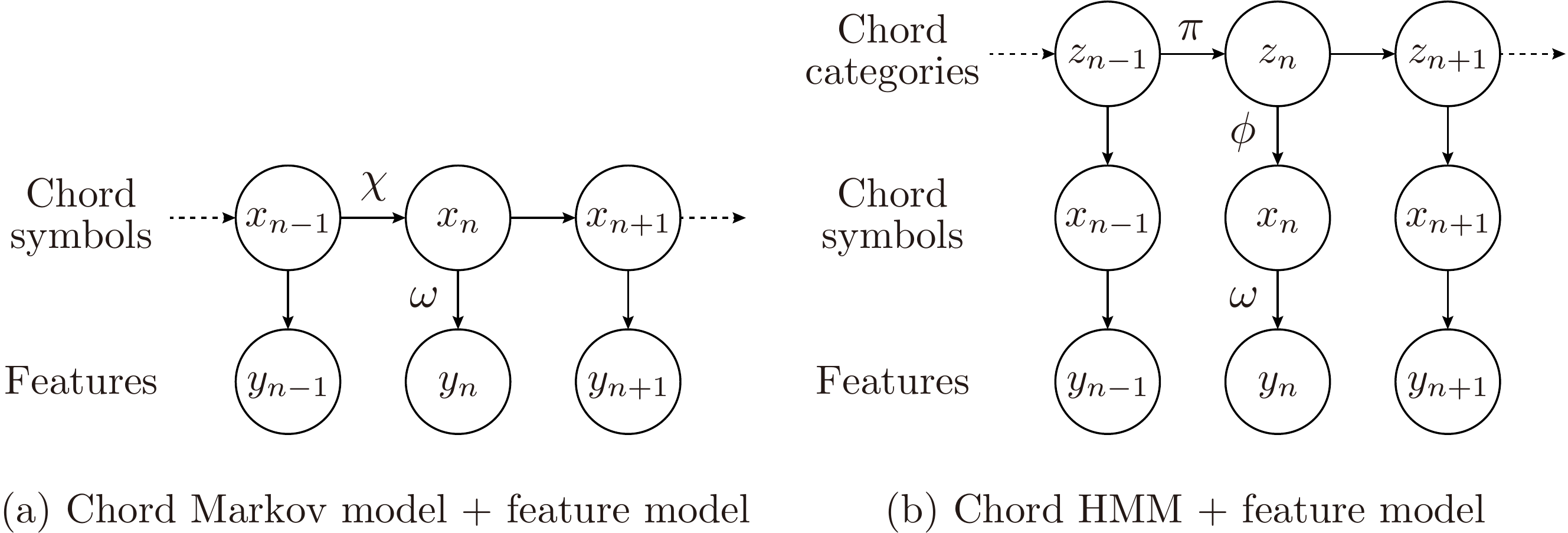}
\vspace{-5mm}
\end{center}
\caption{Integration of chord models with feature models for music processing.}
\label{fig:HMMForApplications}
\vspace{-3mm}
\end{figure*}
As is confirmed for speech recognition, the perplexity of the language model in a recognition system is directly related to the recognition accuracy \cite{Chen1999}.
In our case, the proposed models are expected to improve accuracies for chord recognition and melody harmonisation.
In these applications, the chord model is usually combined with an acoustic model or a melody model, which are described as output probabilities $\omega_{xy}=P(y_n=y\,|\,x_n=x)$ of some feature $y$ (e.g.\ acoustic features or melodic notes) conditioned on a chord symbol.
If one uses Markov models as the chord model, the total model becomes an HMM.
If one uses our HMMs as the chord model, the total model is still an HMM but with an additional layer of latent variables (Fig.~\ref{fig:HMMForApplications}).
To obtain the optimal chord sequence given the observed features $\bm y=y_{1:N}$, the following probability is used:
\begin{equation}
P(\bm x|\bm y)=\frac{P(\bm x,\bm y)}{P(\bm y)}=\frac{\sum\limits_{\bm z}P(\bm z,\bm x,\bm y)}{P(\bm y)}.
\end{equation}
This probability can be computed by GS or other sampling methods.
The situation is similar when one uses our PCFG models as the chord model.
Application for melody harmonisation using the PCFG models extended with a rhythmic model has been studied in Ref.~\cite{Tsushima2017}.

With regard to computational cost, the PCFG models require time complexity of square order with respect to the sequence length and cubic order with respect to the number of nonterminals, which is typically much larger than that of HMMs.
Given the result that PCFG models did not much improve the predictive power of chord sequences compared to HMMs, there would be little interest in using PCFG models for simple recognition tasks.
Nevertheless, they may have importance in music analysis and composition/arrangement tasks \cite{Steedman1984,Rohrmeier2011,Quick2016,Tsushima2017}, where the tree structured grammar is considered useful.

Finally, we discuss implications of our result compared to those in previous studies \cite{Rohrmeier2008,Jacoby2015} that also investigated unsupervised learning of chord categories.
These studies showed that chord categories similar to traditional harmonic functions can be learned from data and our result is a confirmation of this result.
A difference is that whereas in previous studies the use of chord categories was assumed and the model learning was based on comparison between different possibilities of categorisation, in our study we showed that the introduction of chord categories in fact improves the predictive power compared to the case without introducing categories for some data.
Thus our result implies the need for introducing chord categories in harmony models from the informatics viewpoint.

\section{Conclusion}

We have studied self-emergent HMMs and PCFG models as generative statistical models for chord symbols whose latent grammatical structure can be learned unsupervisedly from data.
We have investigated the influence of training-data size and the model size and found that these models outperform Markov models in terms of predictive power particularly for small training-data sizes and small numbers of possible chord symbols, which are typical situations in applications for music processing.
By analysing self-emergent grammars of the models, we have confirmed that they tend to capture syntactic similarities of chords and the learned categories often corresponded to traditional harmonic functions when the model size is small.

These models can be useful as a prior language model of chord sequences for music processing such as chord recognition and melody harmonisation, as they are expected to improve the accuracy compared to conventionally used Markov models.
In addition, as the notion of harmonic functions have been used by musicians to learn harmonic grammar and to create/analyse music, these models can also be useful for automatic composition/arrangement and computational music analysis.
For example, the learned model could indicate users for possibilities for chord substitutions in the process of creating chord progressions.

It is certainly interesting to apply the models for other data in different styles and to compare their syntactic structures.
Because it is based on unsupervised learning, the proposed modelling framework can be even more useful for uncommon music styles for which data exists but no expert knowledge is developed.
It would lead to the development of style-specific harmony theories and quantitative and objective comparisons among them.
This would open a new way of data-driven computational musicology based on the information-scientific aspect of data.

\section*{Acknowledgment}

E. Nakamura is supported by the JSPS Postdoctoral Research Fellowship. This work is in part supported by JSPS KAKENHI Nos.\ 24220006, 26280089, 26700020, 15K16054, 16H01744, 16H02917, 16K00501, and 16J05486 and JST ACCEL No.\ JPMJAC1602.

\appendix
\section{Embedding of HMMs in PCFG Models}
\label{app:HMMEmbedding}

In this appendix, we explain a strict embedding of HMMs in PCFG models, which is mentioned in Sec.~\ref{sec:ModelRelations}.
The result here validates the approximate embedding as in Eqs.~(\ref{eq:PCFGMimickingHMM1})--(\ref{eq:PCFGMimickingHMM4}) and explains why we use such approximations in our study.

To show the embedding relation, we reformulate HMMs.
Instead of Eq.~(\ref{eq:dataSeq}), we represent a data sequence in the following form:
\begin{equation}
x_1\cdots x_N\,{\sf end}~.
\end{equation}
That is, we formally add the symbol `{\sf end}' at the end of each sequence.
As a model that generates this type of data, the state space of an HMM is now extended to $\Gamma'=\Gamma\cup\{\underline{\sf end}\}$, where the added state `$\underline{\sf end}$' only outputs `{\sf end}'.
The initial and output probabilities are defined without any changes, but the transition probabilities are extended as a function $\pi':\Gamma\times\Gamma'\to\mathbb{R}_{\geq0}$ satisfying
\begin{equation}
1=\sum_{w\in\Gamma}\pi'_{zw}+\pi'_{z\,\underline{\sf end}}\quad(^\forall z\in\Gamma).
\label{eq:extendedTrProb}
\end{equation}
The difference to the HMMs used in the main part is the existence of $\pi'_{z\,\underline{\sf end}}$, which controls the expected length of generated sequences, and the following normalisation condition holds for this new HMM:
\begin{equation}
1=\sum_{N=1}^\infty\sum_{\substack{x_1,\ldots,x_N\in\Omega\\z_1,\ldots,z_N\in\Gamma}}
P_{\rm HMM}(x_{1:N}\,{\sf end},z_{1:N}).
\end{equation}
This can be easily derived from Eqs.~(\ref{eq:HMMTrProb}), (\ref{eq:HMMOutProb}), and (\ref{eq:extendedTrProb}).

\begin{figure}[t]
\begin{center}
\includegraphics[clip,width=0.6\columnwidth]{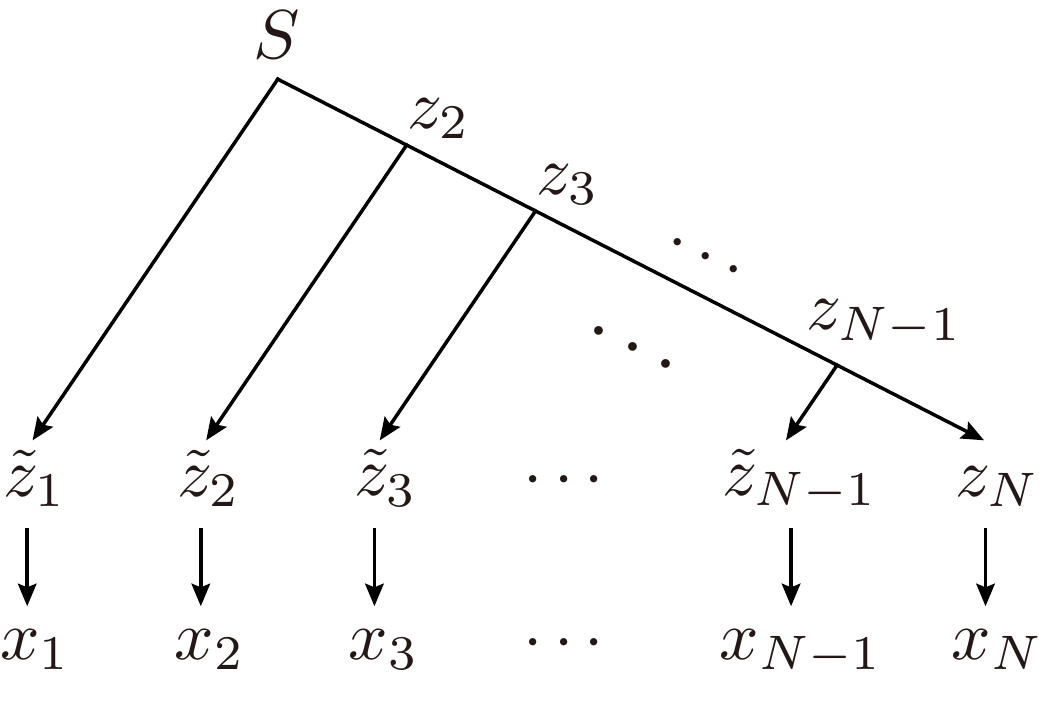}
\vspace{-7mm}
\end{center}
\caption{Derivation tree of a PCFG model that mimics an HMM.}
\label{fig:HMMAsPCFG_strict}
\vspace{-3mm}
\end{figure}

Now let us construct a PCFG model with $2N_\Gamma$ nonterminals (plus the start symbol $S$), where $N_\Gamma=\#\Gamma$ is the state-space size of the HMM.
For each $z\in\Gamma$, we label half of the nonterminals with $z$ and the other half with $\tilde{z}$.
We set the probabilities for this PCFG model as follows (here, $z,w,u\in\Gamma$ and $x\in\Omega$):
\begin{align}
\theta_{S\to\tilde{z}w}&=\pi^{\rm ini}_z\pi'_{zw},\quad
\psi_{S\to x}=\sum_{z\in\Gamma}\pi^{\rm ini}_z\pi'_{z\,\underline{\sf end}}\phi_{zx},
\label{eq:PCFGRealMimickingHMM1}
\\
\theta_{z\to \tilde{u}w}&=\delta_{zu}\pi'_{zw},\quad~
\psi_{z\to x}=\pi'_{z\,\underline{\sf end}}\phi_{zx},
\label{eq:PCFGRealMimickingHMM2}
\\
\psi_{\tilde{z}\to x}&=\phi_{zx}, 
\label{eq:PCFGRealMimickingHMM3}
\end{align}
and other components that are not displayed are all zero.
It can be easily checked that this PCFG model only generates derivation trees of the form in Fig.~\ref{fig:HMMAsPCFG_strict} and has the evidence probability $P_{\rm PCFG}(\bm x)$ equal to that of the HMM.
That is, for any positive integer $N$ and any sequence $x_{1:N}$, we have
\begin{equation}
P_{\rm HMM}(x_{1:N}\,{\sf end})=P_{\rm PCFG}(x_{1:N}),
\end{equation}
which is what we wanted to obtain.

The resemblance between Eqs.(\ref{eq:PCFGMimickingHMM1})--(\ref{eq:PCFGMimickingHMM4}) and Eqs.(\ref{eq:PCFGRealMimickingHMM1}) and (\ref{eq:PCFGRealMimickingHMM2}) is manifest, where the probability $\pi'_{z\,\underline{\sf end}}$ is simplified as $\kappa$ and $\kappa'$ in the former case.
Notice that we needed twice as many nonterminals to constrain the generated derivation trees to those equivalent to the linear-chain grammar.
This significantly increases the computational cost for learning and inference as we discussed in Sec.~\ref{sec:Discussion}.
On the other hand, as we argued in Sec.~\ref{sec:ModelRelations}, when we initialise PCFG model parameters by HMM parameters, it is necessary to relax strict constraints to allow model parameters to explore a wider space for learning.
This is the practical reason why we used the approximate relation for initialising PCFG models.

\section{Normalised Evidence Probability for PCFG Models}
\label{app:NormalisedEvidenceProb}

As mentioned in Sec.~\ref{sec:EvaluationMeasure}, we need to appropriately normalise the evidence probability $P(X)$ for PCFG models to define the test-data perplexity.
Here, we consider the case of a single sequence $\bm x$, which is to be applied for each sequence in the test data.
The evidence probability for PCFG models is normalised as
\begin{equation}
\sum_{\bm y\in\Omega^*}P(\bm y)=1,
\end{equation}
where the sum is taken over all sequences of any length ($\Omega^*=\{y_{1:N}|N\geq1\}$).
What we want is an evidence probability $\bar{P}(\bm x)$ that is normalised in the set of sequences of a fixed length, as in the case for Markov models and HMMs.
Let $N(\bm x)$ denote the length of a sequence $\bm x$.
We want
\begin{equation}
\sum_{\bm y\in\Omega^{N(\bm x)}}\bar{P}(\bm y)=1,
\end{equation}
where $\Omega^{N}$ denotes the set of all sequences of length $N$.
Now, let $P(N)$ be the probability that any sequence of length $N$ would be generated, which is given as
\begin{equation}
P(N)=\sum_{\bm y\in\Omega^{N}}P(\bm y).
\end{equation}
Then $\bar{P}(\bm x)$ is given as $P(\bm x)/P(N(\bm x))$.

It remains to calculate $P(N)$ for a given $N$.
We define modified inside variables $\bar{B}_{nm}(z)$, which are inductively defined through the following modified inside algorithm:
\begin{align}
\bar{B}_{nn}(z)&=\sum_{x\in\Omega}\psi_{z\to x},
\\
\bar{B}_{nm}(z)&=\sum_{z_L,z_R\in\Delta}\sum_{k=1}^{m-1}\theta_{z\to z_Lz_R}\bar{B}_{nk}(z_L)\bar{B}_{(k+1)m}(z_R).
\end{align}
We then have $P(N)=\bar{B}_{1N}(S)$.

\section{Formulas for Symbol-Wise Prediction Probabilities}
\label{app:SymbolPrediction}

Here, we derive formulas for the symbol-wise prediction probability $P(x_n=y\,|\bm x_{\neg n})$ in Eq.~(\ref{eq:SymbolPredictionProb}) for Markov models, HMMs, and PCFG models.
Since we consider this probability as a function of $y$, we have
\begin{equation}
P(x_n=y\,|\bm x_{\neg n})\propto P(x_n=y,\bm x_{\neg n}),
\end{equation}
which will be used to derive the following formulas.

For a $k$\,th-order Markov model, we have
\begin{equation}
P(x_n|\bm x_{\neg n})\propto\prod_{n'=n}^{n+k}\chi_{x_{n'-k:n'-1}x_{n'}}
\end{equation}
for $k+1\leq n\leq N-k$, which will be a general formula.
For samples near the beginning or the end of the sequence, we need a modification to this formula.
For example, for a first-order Markov model,
\begin{align*}
&P(x_n=y\,|\bm x_{\neg n})\propto\chi_{x_{n-1}y}\chi_{yx_{n+1}}\quad (2\leq n\leq N-1);
\\
&P(x_1=y\,|\bm x_{\neg 1})\propto\chi^{\rm ini}_y\chi_{yx_2};
\\
&P(x_N=y\,|\bm x_{\neg N})\propto\chi_{x_{N-1}y},
\end{align*}
and for a second-order Markov model,
\begin{align*}
&P(x_n=y\,|\bm x_{\neg n})\propto\chi_{x_{n-2}x_{n-1}y}\chi_{x_{n-1}yx_{n+1}}\chi_{yx_{n+1}x_{n+2}}
\\
&\hspace{150pt} (3\leq n\leq N-2);
\\
&P(x_1=y\,|\bm x_{\neg 1})\propto\chi^{\rm ini}_y\chi^{{\rm ini}(2)}_{yx_2}\chi_{yx_2x_3};
\\
&P(x_2=y\,|\bm x_{\neg 2})\propto\chi^{{\rm ini}(2)}_{x_1y}\chi_{x_1yx_3}\chi_{yx_3x_4};
\\
&P(x_{N-1}=y\,|\bm x_{\neg (N-1)})\propto\chi_{x_{N-3}x_{N-2}y}\chi_{x_{N-2}yx_N};
\\
&P(x_N=y\,|\bm x_{\neg N})\propto\chi_{x_{N-2}x_{N-1}y}.
\end{align*}
The formulas for higher-order models are similar.

For HMMs, let us notate the forward and backward variables as
\begin{equation}
\alpha_n(z_n)=P(z_n,x_{1:n}),\quad \beta_n(z_n)=P(x_{n+1:N}|z_n),
\end{equation}
which are obtained by the forward and backward algorithms \cite{Rabiner1989}.
With the Markovian assumptions, we have
\begin{align}
&P(z_{n-1},z_n,\bm x)=P(z_{n-1},z_n,\bm x_{\neg n})P(x_n|z_n),
\\
&P(z_{n-1},z_n,\bm x_{\neg n})
\notag
\\
&\quad=P(z_{n-1},x_{1:n-1})P(z_n|z_{n-1})P(x_{n+1:N}|z_n),
\end{align}
and hence the following formula:
\begin{align*}
P(x_n=y\,|\bm x_{\neg n})&\propto\sum_{z_{n-1},z_n\in\Delta}P(z_{n-1},z_n,\bm x)\bigg|_{x_n=y}
\\
&=\!\!\!\!\sum_{z_{n-1},z_n\in\Delta}\!\!\!\!\pi_{x_{n-1}y}\phi_{z_ny}\alpha_{n-1}(z_{n-1})\beta_n(z_n)
\end{align*}
for $n>1$.
The case for $n=1$ is similar.

For PCFG models, the inside and outside variables are defined as
\begin{align}
B_{nm}(z)&=P(z\Rightarrow x_{n:m}),
\\
A_{nm}(z)&=P(S\Rightarrow x_{1:n-1}\,z\,x_{m+1:N}),
\end{align}
where the symbol `$\Rightarrow$' means that the right symbol/sequence is derived from the left symbol/sequence.
These variables can be computed by the inside and outside algorithms \cite{Manning}.
Especially, we have
\begin{equation}
A_{nn}(z)=P(S\Rightarrow x_{1:n-1}\,z\,x_{n+1:N})
\end{equation}
and hence the following formula:
\begin{equation}
P(x_n=y\,|\bm x_{\neg n})\propto\sum_{z\in\Delta}\psi_{z\to y}A_{nn}(z).
\end{equation}


\begin{thebibliography}{99}\setlength\itemsep{-2pt}

\bibitem{Rameau}
J.-P. Rameau, {\em Treatise on Harmony}, Dover Publications, 1971.
\newblock (The original French version was first published in 1722).

\bibitem{Riemann}
H.~Riemann, {\em Harmony Simplified: Or the Theory of the Tonal Functions of
  Chords (2nd ed.)}, Augener, 1896.

\bibitem{Schoenberg}
A.~{Sch\"onberg}, {\em Structural Functions of Harmony (revised ed.)}, W.\ W.\
  Norton \& Company, 1969.

\bibitem{Maler}
W.~Maler, {\em Beitrag zur Durmolltonalen Harmonielehre I (7th ed.)}, F.\ E.\
  C.\ Leuckart, 2007.

\bibitem{TonalHarmony}
S.~Kostka, D.~Payne, and B.~{Alm\'en}, {\em Tonal Harmony with an Introduction
  to Twentieth-Century Music (7th ed.)}, McGraw-Hill, 2013.

\bibitem{BerkleeMethod}
J.~Mulholland and T.~Hojnacki, {\em The Berklee Book of Jazz Harmony}, Berklee
  Press, 2013.

\bibitem{Papadopoulos2007}
H.~Papadopoulos and G.~Peeters, ``Large-Scale Study of Chord Estimation
  Algorithms Based on Chroma Representation and HMM,'' in {\em Proc.\ CBMI},
  pp.\ 53--60, 2007.

\bibitem{Mauch2008}
M.~Mauch and S.~Dixon, ``A Discrete Mixture Model for Chord Labelling,'' in
  {\em Proc.\ ISMIR}, pp.\ 45--50, 2008.

\bibitem{Scholz2009}
R.~Scholz, E.~Vincent, and F.~Bimbot, ``Robust Modeling of Musical Chord
  Sequences Using Probabilistic N-Grams,'' in {\em Proc.\ ICASSP}, pp.\ 53--56,
  2009.

\bibitem{Lewandowski2013}
N.~Boulanger-Lewandowski, Y.~Bengio, and P.~Vincent, ``Audio Chord Recognition
  with Recurrent Neural Networks,'' in {\em Proc.\ ISMIR}, pp.\ 335--340, 2013.

\bibitem{McVicar2014}
M.~McVicar, R.~Santos-Rodr\'iguez, Y.~Ni, and T.~De Bie, ``Automatic Chord
  Estimation from Audio: A Review of the State of the Art,'' {\em IEEE/ACM
  TASLP}, {\bf 22}{\bf(2)}, pp.~556--575, 2014.

\bibitem{Paiement2006}
J.~F. Paiement, D.~Eck, and S.~Bengio, ``Probabilistic Melodic Harmonization,''
  in {\em Proc.\ Conf. Canadian Soc.\ for Comp.\ Studies of Intelligence}, pp.\
  218--229, 2006.

\bibitem{Morris2008}
D.~Morris, I.~Simon, and S.~Basu, ``MySong: Automatic Accompaniment Generation
  for Vocal Melodies,'' in {\em Proc.\ CHI}, pp.\ 725--734, 2008.

\bibitem{Raczynski2013}
S.~A. Raczynski, S.~Fukayama, and E.~Vincent, ``Melody Harmonization with
  Interpolated Probabilistic Models,'' {\em J.\ New Music Res.}, {\bf
  42}{\bf(3)}, pp.~223--235, 2013.

\bibitem{Groves2013}
R.~Groves, ``Automatic Harmonization Using a Hidden Semi-Markov Model,'' in
  {\em Proc.\ AIIDE}, pp.\ 48--54, 2013.

\bibitem{Harmonisation}
M.~Allan and C.~Williams, ``Harmonising Chorales by Probabilistic Inference,''
  in {\em Proc.\ NIPS}, pp.\ 25--32, 2005.

\bibitem{Whorley2016}
R.~P. Whorley and D.~Conklin, ``Music Generation from Statistical Models of
  Harmony,'' {\em J.\ New Music Res.}, {\bf 45}{\bf(2)}, pp.~160--183, 2016.

\bibitem{Winograd1968}
T.~Winograd, ``Linguistics and the Computer Analysis of Tonal Harmony,'' {\em
  J.\ Music Theory}, {\bf 12}{\bf(1)}, pp.~2--49, 1968.

\bibitem{Steedman1984}
M.~J. Steedman, ``A Generative Grammar for Jazz Chord Sequences,'' {\em Music
  Perception}, {\bf 2}{\bf(1)}, pp.~52--77, 1984.

\bibitem{Raphael2004}
C.~Raphael and J.~Stoddard, ``Functional Harmonic Analysis Using Probabilistic
  Models,'' {\em Comp.\ Music\ J.}, {\bf 28}{\bf(3)}, pp.~45--52, 2004.

\bibitem{Temperley2009}
D.~Temperley, ``A Unified Probabilistic Model for Polyphonic Music Analysis,''
  {\em J.\ New Music Res.}, {\bf 38}{\bf(1)}, pp.~3--18, 2009.

\bibitem{Kaneko2010}
H.~Kaneko, D.~Kawakami, and S.~Sagayama, ``Functional Harmony Annotation
  Database for Statistical Music Analysis,'' in {\em Proc.\ ISMIR}, 2010.

\bibitem{Rohrmeier2011}
M.~Rohrmeier, ``Towards a Generative Syntax of Tonal Harmony,'' {\em J.\ Math.\
  Music}, {\bf 5}{\bf(1)}, pp.~35--53, 2011.

\bibitem{Granroth2012}
W.~Granroth and M.~Steedman, ``Statistical Parsing for Harmonic Analysis of
  Jazz Chord Sequences,'' in {\em Proc.\ ICMC}, pp.\ 478--485, 2012.

\bibitem{Fukayama2013}
S.~Fukayama, K.~Yoshii, and M.~Goto, ``Chord-Sequence-Factory: A Chord
  Arrangement System Modifying Factorized Chord Sequence Probabilities,'' in
  {\em Proc.\ ISMIR}, pp.\ 457--462, 2013.

\bibitem{Quick2015}
D.~Quick, ``Composing with Kulitta,'' in {\em Proc.\ ICMC}, pp.\ 306--309,
  2015.

\bibitem{Quick2016}
D.~Quick, ``Learning Production Probabilities for Musical Grammars,'' {\em J.\
  New Music Res.}, {\bf 45}{\bf(4)}, pp.~295--313, 2016.

\bibitem{Conklin2016}
D.~Conklin, ``Chord Sequence Generation with Semiotic Patterns,'' {\em J.\
  Math.\ Music}, {\bf 10}{\bf(2)}, pp.~92--106, 2016.

\bibitem{Manning}
C.~Manning and H.~{Sch\"utze}, {\em Foundations of Statistical Natural Language
  Processing}, MIT Press, Cambridge, 1999.

\bibitem{Rabiner1989}
L.~Rabiner, ``A Tutorial on Hidden Markov Models and Selected Applications in
  Speech Recognition,'' {\em Proc.\ IEEE}, {\bf 77}{\bf(2)}, pp.~257--286,
  1989.

\bibitem{Chen1999}
S.~F. Chen and J.~Goodman, ``An Empirical Study of Smoothing Techniques for
  Language Modeling,'' {\em Computer Speech and Language}, {\bf 13},
  pp.~359--394, 1999.

\bibitem{Bishop}
C.~M. Bishop, {\em Pattern Recognition and Machine Learning}, Springer, 2006.

\bibitem{BeatlesData}
C.~Harte, M.~B. Sandler, S.~A. Abdallah, and E.~{G\'omez}, ``Symbolic
  Representation of Musical Chords: A Proposed Syntax for Text Annotations,''
  in {\em Proc.\ ISMIR}, pp.\ 66--71, 2005.

\bibitem{Yoshii2011}
K.~Yoshii and M.~Goto, ``A Vocabulary-Free Infinity-Gram Model for
  Nonparametric Bayesian Chord Progression Analysis,'' in {\em Proc.\ ISMIR},
  pp.\ 645--650, 2011.

\bibitem{Johnson2007}
M.~Johnson, ``Why Doesn't EM Find Good HMM POS-Taggers?,'' in {\em Proc.\
  EMNLP-CoNLL}, pp.\ 296--305, 2007.

\bibitem{Goldwater2007}
S.~Goldwater and T.~Griffiths, ``A Fully Bayesian Approach to Unsupervised
  Part-of-Speech Tagging,'' in {\em Proc.\ ACL}, pp.\ 744--751, 2007.

\bibitem{Johnson1998}
M.~Johnson, ``PCFG Models of Linguistic Tree Representations,'' {\em
  Computational Linguistics}, {\bf 24}, pp.~613--632, 1998.

\bibitem{Matsuzaki2005}
T.~Matsuzaki, Y.~Miyao, and J.~Tsujii, ``Probabilistic CFG with Latent
  Annotations,'' in {\em Proc.\ ACL}, pp.\ 75--82, 2005.

\bibitem{Johnson}
M.~Johnson, T.~Griffiths, and S.~Goldwater, ``Bayesian Inference for PCFGs via
  Markov Chain Monte Carlo,'' in {\em Proc.\ HLT-NAACL}, pp.\ 139--146, 2007.

\bibitem{GTTM}
F.~Lerdahl and R.~Jackendoff, {\em A Generative Theory of Tonal Music}, MIT
  Press, Cambridge, 1983.

\bibitem{Chomsky}
N.~Chomsky, {\em Syntactic Structures}, Mouton \& Co., 1957.

\bibitem{Rohrmeier2008}
M.~Rohrmeier and I.~Cross, ``Statistical Properties of Tonal Harmony in Bach's
  Chorales,'' in {\em Proc.\ ICMPC}, pp.\ 619--627, 2008.

\bibitem{Jacoby2015}
N.~Jacoby, N.~Tishby, and D.~Tymoczko, ``An Information Theoretic Approach to
  Chord Categorization and Functional Harmony,'' {\em J.\ New Music Res.}, {\bf
  44}{\bf(3)}, pp.~219--244, 2015.

\bibitem{Mavromatis2009}
P.~Mavromatis, ``Minimum Description Length Modelling of Musical Structure,''
  {\em J.\ Math.\ Music}, {\bf 3}{\bf(3)}, pp.~117--136, 2009.

\bibitem{Mavromatis2012}
P.~Mavromatis, ``Exploring the Rhythm of the Palestrina Style: A Case Study in
  Probabilistic Grammar Induction,'' {\em J.\ Music Theory}, {\bf 56}{\bf(2)},
  pp.~169--223, 2012.

\bibitem{Abdallah2014}
S.~A. Abdallah and N.~E. Gold, ``Comparing Models of Symbolic Music Using
  Probabilistic Grammars And Probabilistic Programming,'' in {\em
  Proc.~ICMC/SMC}, pp.\ 1524--1531, 2014.

\bibitem{Lari1990}
K.~Lari and S.~J. Young, ``The Estimation of Stochastic Context-Free Grammars
  Using the Inside-Outside Algorithm,'' {\em Computer speech \& language}, {\bf
  4}{\bf(1)}, pp.~35--56, 1990.

\bibitem{Burgoyne2011}
J.~A. Burgoyne, J.~Wild, and I.~Fujinaga, ``An Expert Ground Truth Set for
  Audio Chord Recognition and Music Analysis,'' in {\em Proc.\ ISMIR}, pp.\
  633--638, 2011.

\bibitem{Tsushima2017}
H.~Tsushima, E.~Nakamura, K.~Itoyama, and K.~Yoshii, ``Function- and
  Rhythm-Aware Melody Harmonization Based on Tree-Structured Parsing and
  Split-Merge Sampling of Chord Sequences,'' in {\em Proc.\ ISMIR}, pp.\
  502--508, 2017.

\end{thebibliography}
\end{document}